\documentclass[runningheads]{llncs}

\usepackage{eccv}

\usepackage{eccvabbrv}

\usepackage{graphicx}
\usepackage{booktabs}

\usepackage[accsupp]{axessibility}  % Improves PDF readability for those with disabilities.

\usepackage{hyperref}

\usepackage{orcidlink}

\usepackage{algorithm}
\usepackage{algorithmic}
\usepackage{multirow}
\usepackage{multicol}
\usepackage{tablefootnote}
\usepackage{subfiles}

\begin{document}

\title{Random Walk on Pixel Manifolds for Anomaly Segmentation of Complex Driving Scenes} 

\titlerunning{RWPM}

\author{Zelong Zeng\inst{1}\orcidlink{0000-0003-3740-2743} \and
Kaname Tomite \inst{1}}

\authorrunning{Z.~Zeng et al.}

\institute{SenseTime Japan, Tokyo, Japan\\
\email{\{zengzelong,tomite\}@sensetime.jp}}

\maketitle

\begin{abstract}
    In anomaly segmentation for complex driving scenes, state-of-the-art approaches utilize anomaly scoring functions to calculate anomaly scores. 
    For these functions, accurately predicting the logits of inlier classes for each pixel is crucial for precisely inferring the anomaly score. 
    However, in real-world driving scenarios, the diversity of scenes often results in distorted manifolds of pixel embeddings in the space. This effect is not conducive to directly using the pixel embeddings for the logit prediction during inference, a concern overlooked by existing methods. 
    To address this problem, we propose a novel method called Random Walk on Pixel Manifolds (RWPM). RWPM utilizes random walks to reveal the intrinsic relationships among pixels to refine the pixel embeddings. 
    The refined pixel embeddings alleviate the distortion of manifolds, improving the accuracy of anomaly scores. 
    Our extensive experiments show that RWPM consistently improve the performance of the existing anomaly segmentation methods and achieve the best results. 
    Code is available at: \url{https://github.com/ZelongZeng/RWPM}.
    
\end{abstract}

\section{Introduction}
\label{sec:intro}
Semantic segmentation~\cite{chen2018encoder,cheng2021per,xie2021segformer,cheng2022masked,cordts2016cityscapes} plays a crucial role in the field of automated driving, which can aid the automated driving system in perceiving and understanding the surrounding environment, thereby facilitating the system make accurate decision. 
Typically, the semantic segmentation models are trained on closed-datasets with fixed semantic categories, showing good recognition capabilities for inlier classes (\eg, roads, vehicles, etc.). 
However, these models often fail to effectively detect potential outlier class objects (\eg, animals, fallen tires, etc.) in real-world scenarios. 
Due to the critical importance of correctly detecting outlier class targets for the safety of autonomous driving, there is a growing focus on addressing this task, namely, anomaly segmentation. 

The standard paradigm for anomaly segmentation is assigning an anomaly score to each pixel of an input image, where a higher score indicates a greater probability that the pixel belongs to an outlier. Thus, the way used for estimating the anomaly score is crucial for anomaly segmentation methods. 
In early approaches, the anomaly score is often given by estimating the prediction uncertainty~\cite{gal2016dropout,lakshminarayanan2017simple,hendrycks2017a,liang2018enhancing}, or by comparing the discrepancy between the original image and its reconstructed image generated based on the semantic segmentation predictions~\cite{creusot2015real,lis2019detecting,ohgushi2020road,di2021pixel}. 
% However, due to the neglect of constraints, these methods are prone to detect false anomalies~\cite{di2021pixel,tian2022pixel}. 
Different from early methods, recent methods~\cite{tian2022pixel,choi2023balanced,liu2023residual,nayal2023rba,rai2023unmasking,grcic2023advantages} adopt Outlier Exposure (OE) strategy~\cite{hendrycks2018deep}, which directly learn a pixel-level anomaly class to detect anomalous pixels that deviate from inlier classes. 
% yielding better performance compared to previous approaches. 
Specifically, these methods propose anomaly scoring functions to infer the anomaly scores and, based on the functions, leverage auxiliary outlier data to fine-tune a semantic segmentation model. 
The auxiliary outlier data is typically generated by extracting outlier-class objects from other datasets (\eg, COCO~\cite{lin2014microsoft}) and incorporating them into driving scene images~\cite{tian2022pixel,nayal2023rba,rai2023unmasking}. 
Furthermore, some of them~\cite{nayal2023rba,rai2023unmasking,grcic2023advantages} adopt mask-based segmentation networks instead of pixel-based segmentation networks, achieving state-of-the-art (SOTA) performance in anomaly segmentation. 
In this paper, we concentrate on the current methods using OE strategy. These methods indicate superior performance to the early methods using prediction uncertainty or image reconstruction. 

\begin{figure}[tb]
  \centering
  \includegraphics[width=0.9\textwidth]{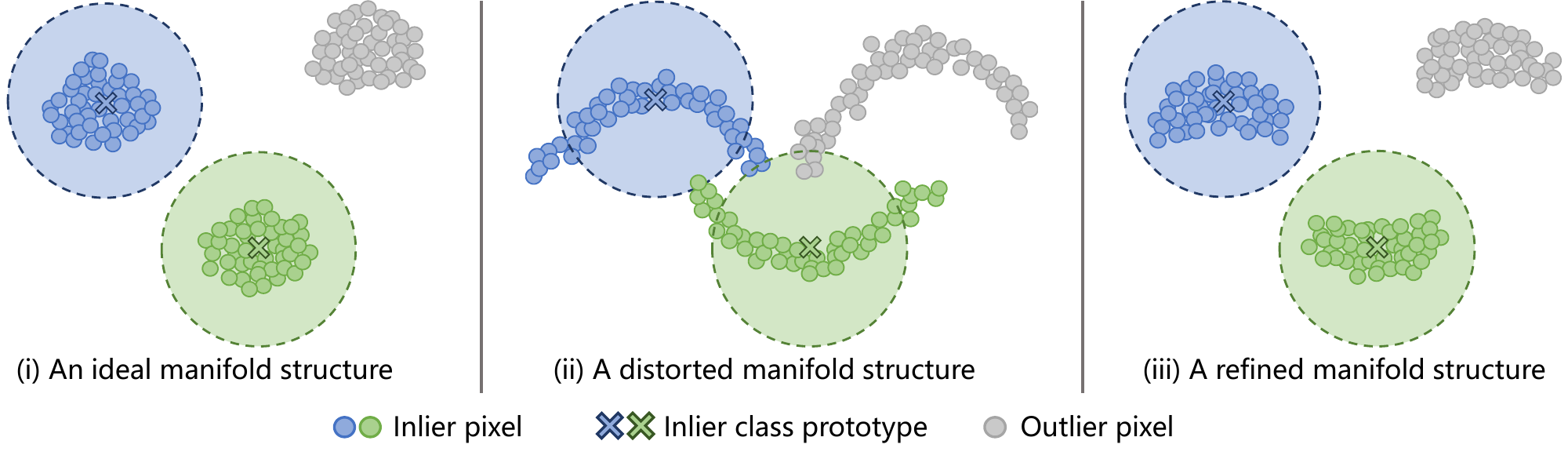}
  \caption{\textbf{A toy example about manifolds of pixel embeddings.} The dashed circles indicate regions with high predicted logit to the corresponding inlier class. For the pixel in these regions, its anomaly score is small. The prototype is the vector of network weights of the classifier for the corresponding inlier class.  
  (i) An ideal manifold structure. All inlier pixels are within the corresponding high logit region, whereas all outlier pixels are outside the areas. The anomaly score derived from the anomaly scoring function effectively discriminates between them. 
  (ii) The manifolds are affected by the diversity of data, causing some inlier pixels to deviate from the corresponding high logit region, while some outlier pixels approach the regions. This results in false positives/negatives in the anomaly scores. \textbf{However, pixels of the same class are still in the same manifold, indicating that the manifold structure can be utilized to reveal the intrinsic relationships between pixels.}
  (iii) Our RWPM utilizes random walks to capture the manifold structure to diffuse and update pixel embeddings. The embeddings within the same manifold tend to become more similar after the updating. 
  \textbf{Fig~\ref{fig:visualization} and Fig~\ref{fig:qualitative} respectively present the visualization of embedding distributions and the qualitative results of real examples, consistent with the description of this toy example.}}
  \label{fig:manifolds1}
\end{figure}

The anomaly scoring functions in OE methods infer anomaly scores based on the inlier class logit predictions from a semantic segmentation model. 
Specifically, the logic of these functions can be represented as follows: 
when the predicted logits of a pixel embedding are small across all inlier classes, indicating that the pixel is dissimilar to all inlier classes and is an outlier with high probability, the function yields a higher anomaly score. 
Therefore, accurately measuring the logits of each pixel is crucial for precisely evaluating the anomaly score. 
However, in practical open-world driving scenarios, various environmental conditions (such as lighting, road materials, etc.) and a diverse range of unknown class objects are present. 
During the inference phase, the diversity of driving scenes often results in distorted manifolds of pixel embedding in the embedding space, causing pixel embeddings to deviate from the ideal distribution. 
This phenomenon has been validated in previous works of metric learning~\cite{bai2017regularized,iscen2017efficient,yang2019efficient,yang2020mining}. 
Since the logits are calculated from the inner product (\ie, cosine distance) of the pixel embeddings and the vectors of classifiers' network weights, the distorted manifolds are not conducive to directly using the pixel embeddings for the logit prediction. 
However, current anomaly segmentation methods ignore this issue and directly use the pixel embeddings for the logit prediction in the inference phase, so the results often fail to accurately reflect the anomaly score for the corresponding pixels, causing high false positive/negative rate (see Fig.~\ref{fig:manifolds1}(i) and Fig.~\ref{fig:manifolds1}(ii)). 

In this paper, we propose a novel method, called Random Walk on Pixel Manifolds (RWPM), to alleviate the effect of data diversity on the manifolds. 
RWPM utilizes random walks to capture the manifold structure of the pixel embeddings to diffuse and update the embedding of each pixel. 
Since the structure of data manifolds implicitly contains the intrinsic relationship among data points (see Fig.~\ref{fig:manifolds1}(ii))~\cite{zhou2003ranking,bai2017regularized,iscen2018mining,yang2019efficient}, using random walks to measure the similarity between pixels on manifolds (\ie, distance on manifolds rather than Euclidean distance) better reflects the intrinsic relationships between pixels. 
Consequently, diffusing and updating pixel embeddings based on the similarity on manifolds results in high similarity among pixels on the same manifold and low similarity among pixels on different manifolds. 
In the embedding space, this manifests as pixel embeddings on the same manifold forming more compact clusters, thereby mitigating the distortion effect of data diversity on the manifolds, which facilitates the logit prediction (see Fig.~\ref{fig:manifolds1}(iii)). 
Furthermore, since the typically large number of pixels in an image, utilizing all pixels for the random walks results in significant memory and computational overhead, which damages the deployment and real-time execution of our RWPM. 
To address this issue, we introduce the Partial Random Walk strategy for RWPM, reducing memory consumption and enhancing the operational efficiency of RWPM. 
\textit{Note that, RWPM is proposed to use in the inference phase, thus no any additional training is required. } Additionally, RWPM can be directly integrated into existing anomaly segmentation framework without the need for any extra modification to the network structure. 
Extensive experiments in various anomaly segmentation benchmarks of road scenes, namely Fishyscapes~\cite{blum2021fishyscapes}, Road Anomaly~\cite{lis2019detecting}, and SMIYC~\cite{chan2021segmentmeifyoucan}, demonstrate that RWPM consistently improves the performance of existing methods, achieving SOTA results. 
To summarise, our contributions are the following:
\begin{itemize}
  \item We introduce that the diversity of driving scenes results in distorted manifolds of pixel embedding, thereby affecting the accuracy of anomaly scoring functions in inferring anomaly scores. This problem is ignored by existing anomaly segmentation methods. 
  \item We propose the Random Walk on Pixel Manifolds (RWPM) that utilizes random walks to reveal the intrinsic relationships among pixels to refine the pixel embeddings. This way alleviate the effect on manifolds caused by the driving scene diversity. 
  \item We propose the Partial Random Walk strategy for RWPM to reduce the memory consumption and improve the operational efficiency. 
\end{itemize}

\section{Related Work}
\label{sec:realted}

% \subsection{Anomaly Segmentation}
\textbf{Anomaly Segmentation.} 
Anomaly segmentation aims to detect the outlier class object in some specific scenes, such as complex driving scenes~\cite{cordts2016cityscapes,lis2019detecting,chan2021segmentmeifyoucan}. 
Existing anomaly segmentation can be broadly divided into three categories: (a) \textit{Uncertainty-based methods}, (b) \textit{Reconstruction-based methods} and (c) \textit{Outlier Exposure methods}. 
Uncertainty-based methods assume that outlier samples result in low-confidence predictions. 
Based on this assumption, they focus on measuring the pixel-wise anomaly score by estimating the prediction uncertainty through various ways, such as ensembles~\cite{lakshminarayanan2017simple}, Bayesian estimation~\cite{gal2016dropout}, maximum softmax probability~\cite{hendrycks2017a,liang2018enhancing}, and logit~\cite{hendrycks2019scaling,jung2021standardized}. 
Since segmentation models are trained on close-set, they may misclassify unseen classes with high confidence, leading uncertainty-based methods fail to detect anomalies.  
Reconstruction-based methods detect anomaly pixel by comparing the discrepancy between the original image and its reconstructed image generated based on the semantic segmentation predictions~\cite{creusot2015real,lis2019detecting,ohgushi2020road,di2021pixel}. 
These methods are challenged by the accuracy of segmentation prediction and the difficulties of training. 
Recent methods~\cite{chan2021entropy,tian2022pixel,choi2023balanced,rai2023unmasking,nayal2023rba,grcic2023advantages} introduce the Outlier Exposure (OE) strategy~\cite{hendrycks2018deep} that utilizes auxiliary outlier data and anomaly scoring functions to learn models. 
The auxiliary outlier data is the image from other datasets (\eg, COCO~\cite{lin2014microsoft}), or cutting the outlier objects from other datasets and pasting them into the inlier scenes, and the anomaly scoring functions are used to calculate the anomaly score for each pixel. 
Compared with previous works, OE strategy can reach better performance. 
More recently, some OE methods~\cite{rai2023unmasking,nayal2023rba,grcic2023advantages} adopt mask-based segmentation networks~\cite{cheng2021per,cheng2022masked}, instead of pixel-based segmentation networks~\cite{chen2017deeplab,chen2018encoder}, as their backbones. 
Based on the powerful capability of mask-based networks in segmentation tasks, these methods significantly reduces false detection rates, achieving SOTA results. 
In this paper, we only concentrate on the OE methods.  
The anomaly scoring functions employed in OE methods are based on the predicted logits of inlier classes.   
However, during the inference phase, the diversity of driving scenes results in distorted manifolds of pixel embeddings. 
As a result, the scoring functions may fail to use the embeddings to accurately calculate logits, thereby affects the accuracy of measuring anomaly scores. 
Current OE methods ignore this problem. 
Our proposed RWPM addresses this problem by utilizing random walks on pixel manifolds to reveal the intrinsic relationships among pixels and refine the pixel embeddings. 
To the best of our knowledge, RWPM is the first work to propose and resolve this problem.

% \subsection{Random Walk Process}
\noindent \textbf{Mining on Data Manifolds.} 
Mining on data manifolds, also known as diffusion process, is a well-established technique commonly used in retrieval tasks~\cite{zhou2003ranking,bai2017regularized,bai2017ensemble,iscen2017efficient,yang2019efficient}. 
% or semi-supervised learning~\cite{zhou2003learning,iscen2018mining}. 
These works utilize random walks on data manifolds to capture the data manifold structure for revealing the intrinsic relationship between data points. 
The data manifold is constructed by a graph, where each node represents a sample and each edge presents the connection between a node and its neighborhoods, weighted proportionally according to their distance. 
For retrieval tasks, the images of the same object may be shown in different conditions (\eg, lighting and angles). This diversity often leads to the manifold structure is not conducive to accurately ranking using distance metrics directly~\cite{iscen2017efficient,yang2019efficient}. 
Thus, diffusion process is used to diffuse the state (typically initialized to $1$) of a query image on the data manifolds. 
If a sample is on the same data manifold with the query (the distance on manifolds is close), then more energy of the state will flow to that sample. 
Therefore, the final state of all samples can be seen as the accurate ranking scores. 
Moreover, diffusion process is also adopted in semi-supervised learning~\cite{zhou2003learning,iscen2018mining} to propagate the label of labeled samples to unlabeled samples. 
Similarly, some unsupervised segmentation works~\cite{grady2006random,huang2019difference} utilize diffusion process to propagate the labels of manually annotated pixels to the unlabeled pixels. 
These existing works utilize random walks on data manifolds for ranking or label propagation. 
Different from these works, our target is to solve the problem occurring in pixel manifolds for anomaly scoring functions. 
We propose innovatively using random walks to diffuse and update pixel embeddings, eliminating the effect of distorted manifolds directly at the embedding-level, thereby improving the accuracy of subsequent processing. 
 % (segmentation and anomaly score calculation). 

\section{Proposed Method}
\label{sec:method}

\subsection{Preliminaries}
\label{subsec:preliminary_as} 
We denote an image space $\mathcal{X} \subset \mathbb{R}^{3\times H \times W}$, where $H$ and $W$ are the height and width, respectively. 
During the inference phase, the goal for an anomaly segmentation model $f$ is to map an input image $\boldsymbol{x} \in \mathcal{X}$ to an anomaly space, \ie, $f: \boldsymbol{x} \mapsto \mathcal{Z}  \subset \mathbb{R}^{H \times W}$. 
Specifically, an anomaly segmentation model typically consists of the backbone encoder-decoder $F$ and the classifier $C$. 
For an input image $\boldsymbol{x}$, firstly, its pixel embedding map is encoded as $\boldsymbol{p} = F\left(\boldsymbol{x}\right) \in \mathbb{R}^{d \times H \times W}$, where $d$ is the dimension of the pixel embedding. 
Then, its logit map $\boldsymbol{l}$ is predicted by $\boldsymbol{l} = C\left(F\left(\boldsymbol{x}\right)\right) \in \mathbb{R}^{K \times H \times W}$, where $K$ is the number of inlier classes. 
Note that, depending on the backbone network, the size of the embedding map and logit map may sometimes be smaller than the original image. For the sake of convenience in this paper, we do not make a special distinction. 
Finally, a scoring function is used to calculate the anomaly score map $\boldsymbol{z} \in \mathcal{Z}$ based on the logit map $\boldsymbol{l}$. 

For anomaly scoring functions, their logic can be represented as follows: 
when the predicted logits of a pixel embedding are small across all inlier classes, indicating that the pixel is dissimilar to all inlier classes and is likely an outlier, the function generates a higher anomaly score. 
For example, let's define $\boldsymbol{l}_{h,w}\left(k\right)$ is the logit of a pixel $\boldsymbol{x}_{h,w}$ for class $k$. 
The scoring function in PEBAL~\cite{tian2022pixel} is $E_{h,w}=-\log \sum_{k\in\left\{1\cdots K\right\}} \exp\left(\boldsymbol{l}_{h,w}\left(k\right)\right)$. 
Obviously, when the anomaly score $E_{h,w}$ is large, it indicates that the logits of the pixel across all inlier classes are small, \ie, the pixel $\boldsymbol{x}_{h,w}$ is dissimilar from all inlier classes (see Section S.2 of the supplementary material for the analysis of other anomaly scoring functions). 

\begin{figure}[t]
  \centering
  \includegraphics[width=0.95\textwidth]{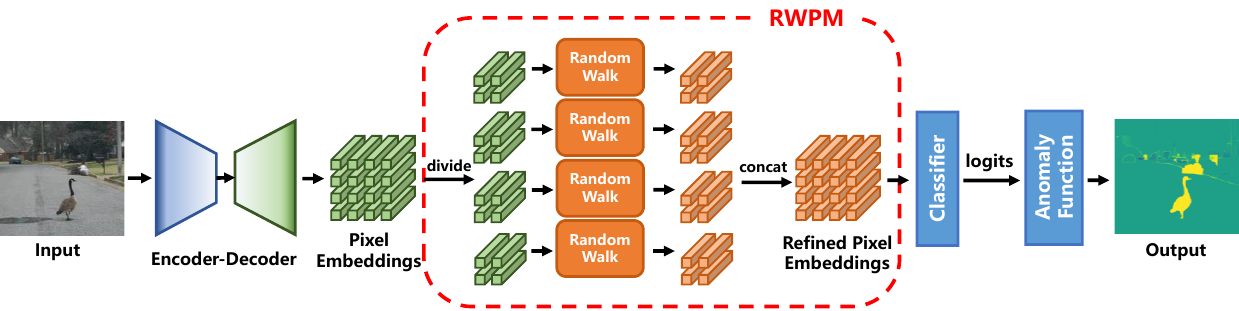}
  \caption{\textbf{Overview.} This figure illustrates the application of RWPM to an existing anomaly segmentation framework during the inference phase. The red dashed box highlights the RWPM part. 
  First, the encoder-decoder is used to extract the pixel embedding map of the input image. 
  The pixel embedding map are subsequently partitioned into $n^2$ sub-maps. 
  For each sub-map, we update its pixel embeddings by using random walks to obtain a refined sub-map. 
  Finally, the refined sub-maps are concatenated to form the refined embedding map, which are then input into the subsequent network structure. 
  Notably, RWPM can be directly integrated into existing frameworks without requiring extra training or changes to the network structure.}
  \label{fig:framework}
\end{figure}

\subsection{Graph Construction}
\label{subsec:graph}
We begin by constructing a graph to represent the manifolds of pixel embeddings. 
In our approach, we utilize an affinity matrix based on cosine similarity to characterize the relationships between pixels. 
Specifically, we first use the encoder-decoder component of an anomaly segmentation model to extract the pixel embeddings $\boldsymbol{p} \in \mathbb{R}^{d \times H \times W}$. 
For ease of description and computation, we reshape the $\boldsymbol{p}$ as $\boldsymbol{p}^{r} \in \mathbb{R}^{ HW \times d}$, where each row represents a pixel's embedding. 
We then normalize each row of $\boldsymbol{p}^{r}$ to obtain $\widehat{\boldsymbol{p}}^{r}$, \ie, $\widehat{\boldsymbol{p}}^{r}_{i} = {\boldsymbol{p}^{r}_{i}}/{\left \| \boldsymbol{p}^{r}_{i} \right \|}$. 
The affinity matrix is define as $\mathbf{W} \in \mathbb{R}^{ HW \times HW}$, where each component is obtained by: 
\begin{equation}
    \mathbf{W}_{ij}=\left\{\begin{matrix} \left \langle \widehat{\boldsymbol{p}}^{r}_{i},\widehat{\boldsymbol{p}}^{r}_{j} \right \rangle & i\neq j\\ 
     0& i=j 
    \end{matrix}\right.
    ,
  \label{eq:affinity}
\end{equation}
where $\left \langle \cdot, \cdot \right \rangle$ denotes inner product, $\mathbf{W}_{ii} = 0$ is guaranteed to circumvent self-loops in the graph represented by $\mathbf{W}$.

To mitigate the influence of noise, locally constrained random walk methods~\cite{iscen2018mining,yang2019efficient,yang2020mining} adopt local constraints to construct the graph, \ie, each point is only connected to its locally nearest neighbors in the graph. 
Determining the nearest neighbors for each point traditionally involves applying nearest neighbor search algorithms, such as the $k$-NN algorithm. 
However, in anomaly segmentation tasks, the size of the graph is large and the graph is constructed online, making the search algorithms impractical. 
To benefit from the local constraints, we utilize a softmax function to normalize each row of $\mathbf{W}$ to obtain locally constrained graph $\mathbf{S}$: 
\begin{equation}
    \mathbf{S}_{ij}=\left\{\begin{matrix}
    \frac{\exp\left ( \mathbf{W}_{ij}/\tau \right )}{\sum_{i=1,i\neq j}^{HW}\exp\left(\mathbf{W}_{ij}/\tau\right)} & i \neq j\\
    0 & i=j
    \end{matrix}\right. 
    ,
  \label{eq:affinity_softmax}
\end{equation}
where $\tau$ is a temperature hyper-parameter, we set its value less than $1.0$ to remove the effect of non-neighbor points. 
Compared to typical search algorithms, Eq.~\ref{eq:affinity_softmax} can impose local constraints more efficiently and effectively (see Section S.4 of the supplementary material), and keep the scale of embedding in subsequent random walks. 

\subsection{Random Walk Process}
\label{subsec:random_walk}
Now we have constructed the graph $\mathbf{S}$, we can start the random walks. 
Different from previous mining on data manifolds works utilizing random walks to propagate the states of known samples to other unknown samples, we use a random walk to propagate the embedding between pixels on manifolds. 

We first define $\mathbf{m}^{t} \in \mathbb{R}^{ HW \times d}$ as the updated pixel embeddings in the $t$-th step of the random walks. $\mathbf{m}^{0}$ is initialized as $\boldsymbol{p}^{r}$. 
Then, each iteration step of the random walks is described as: 
\begin{equation}
    \begin{matrix}
    \mathbf{m}^{t+1} = \alpha \mathbf{S} \mathbf{m}^{t}+\left(1-\alpha\right)\mathbf{m}^{0}, & \alpha \in \left(0,1\right)
    \end{matrix}
    ,
    \label{eq:random_walk}
\end{equation}
where $\alpha$ is the probability continuing the random walk from the current state $\mathbf{m}^{t}$, while $\left(1-\alpha \right)$ represents the probability of restarting from the initial state $\mathbf{m}^{0}$. 
At each iteration, guided by the manifold structure, each pixel diffuses its embedding to other pixels while also receiving embedding from other pixels for updating its own embedding. 
The updated pixel embedding can be seen as a weighted ensemble of other pixel embeddings. If pixels are similar on manifolds, their embeddings will gradually become similar. 
Mathematically speaking, the whole iteration process can be written as a closed-from as follows~\cite{zhou2003ranking}: 
\begin{equation}
    \mathbf{m}^{\infty} = \left(1-\alpha\right)\left(\mathbf{I}-\alpha \mathbf{S}\right)^{-1}\mathbf{m}^{0}
    ,
    \label{eq:random_walk_inf}
\end{equation}
where $\mathbf{I}\in \mathbb{R}^{ HW \times HW}$ is an identity matrix. 
Finally, $\mathbf{m}^{\infty}$ is used as the refined pixel embeddings for the subsequent anomaly segmentation process.  

\subsection{Partial Random Walk}
\label{subsec:prw}
When the size of the image is large, the size of the relevant matrices used for the calculation of random walks becomes extremely huge as well. For instance, when the image size is $\left(512\times1024\right)$, the size of matrix $S$ reaches $\left(524288\times524288\right)$. 
This significantly increases the memory consumption and computational time of RWPM, which is not conducive to online operation. 
Unfortunately, in the field of autonomous driving, the image size of driving scenes is generally large.
To address this problem, we propose Partial Random Walk, which consists of the following two aspects: \textit{embedding map partitioning} and \textit{limited iteration}. 

To improve efficiency, we first propose embedding map partitioning strategy to reduce the size of the pixel embedding map used for random walks. 
Specifically, for a pixel embedding map  $\boldsymbol{p}\in\mathbb{R}^{d \times H \times W}$, we equally divide it into $n\times n$ sub-maps $\left\{\boldsymbol{p}^{(1)},\cdots,\boldsymbol{p}^{(n^2)}\right\}\in\mathbb{R}^{d \times \frac{H}{n} \times \frac{W}{n}}$. 
Then, for each sub-map in $\left\{\boldsymbol{p}^{(1)},\cdots,\boldsymbol{p}^{(n^2)}\right\}$, we follow the steps of Eq.~\ref{eq:affinity} and Eq.~\ref{eq:affinity_softmax} to construct their sub-graphs  $\left\{\boldsymbol{S}_{\boldsymbol{p}^{(1)}}, \cdots, \boldsymbol{S}_{\boldsymbol{p}^{(n^2)}}\right\}\in\mathbb{R}^{d \times \frac{HW}{n^2} \times \frac{HW}{n^2}}$. 
We also attempted other methods to reduce the number of pixel embeddings, such as resize or superpixel~\cite{achanta2012slic}. 
Experimental results indicate that embedding map partitioning is the optimal solution among them (see Section~\ref{sec:ablation}). 

Next, for each sub-map $\boldsymbol{p}^{(i)}$, we perform random walks with the limited iteration strategy in their corresponding sub-graphs $\boldsymbol{S}_{\boldsymbol{p}^{(i)}}$. 
That is, we use Eq.~\ref{eq:random_walk} for a small number of iterations instead of using the closed-form Eq.~\ref{eq:random_walk_inf}. 
We found that a few iterations by using Eq.~\ref{eq:random_walk} are enough to yield high quality results. 
The time complexity of Eq.~\ref{eq:random_walk} with $T$ iterations for each sub-map is $O(Td(\frac{HW}{n^2})^2)$, while the time complexity of Eq.~\ref{eq:random_walk_inf} is $O((\frac{HW}{n^2})^3)+O(d(\frac{HW}{n^2})^2)$. 
Since $Td$ is much smaller than $\frac{HW}{n^2}$, the limited iteration strategy can significantly improve the efficiency of RWPM. 
After the random walks, we obtain the refined pixel embeddings sub-map $\mathbf{m}^{T}_{\boldsymbol{p}^{(i)}}$ for each sub-map $\boldsymbol{p}^{(i)}$. 
Finally, we concatenate the refined sub-maps based on their positions to obtain the final refined pixel embedding map. 
Additionally, When $n$ is large (observed experimentally as $n>2$), due to spatial constraints, the baseline values of anomaly scores outputted on each sub-map may vary. Hence, a calibration is performed. For adjacent refined sub-maps $\boldsymbol{p}^{(i)'}$ and $\boldsymbol{p}^{(j)'}$, we compute the average anomaly scores $I$ and $J$ of all pixels on their adjacent edges, respectively. 
Subsequently, using $\boldsymbol{p}^{(i)'}$ as the reference, we calculate the ratio $\frac{I}{J}$, and then multiply all anomaly scores on sub-map $\boldsymbol{p}^{(j)'}$ by this ratio to achieve the calibration. 

Fig.~\ref{fig:framework} shows the workflow of the proposed RWPM.

\section{Experiment}
\label{sec:experiment}

\noindent \textbf{Dataset:} We evaluate our approach on three standard benchmark datasets for anomaly segmentation, \ie, Fishyscapes Lost\&Found~\cite{blum2021fishyscapes}, Road Anomaly~\cite{lis2019detecting} and Segment Me If You Can (SMIYC)~\cite{chan2021segmentmeifyoucan}. 
Fishyscapes Lost\&Found has 100 validation images. The domain of this dataset is similar to that of Cityscapes~\cite{cordts2016cityscapes} dataset. 
Road Anomaly dataset contains 60 images of real-world road scenes. Unlike the Fishyscapes Lost\&Found dataset, the images in Road Anomaly contain various classes and sizes of anomaly objects, making it more challenging. 
SMIYC consists of two subsets, Anomaly track and Obstacle track. 
The Anomaly track focuses on detecting large anomalous objects, while the Obstacle track emphasizes detecting small anomalous objects on the road.
Anomaly track contains 10 validation images and 100 online test images, while Obstacle track contains 30 validation images and 327 online test images. 
The Road Anomaly and SMIYC datasets both exhibit high domain shifts compared to the Cityscapes dataset. 

\noindent \textbf{Evaluation Metrics: } Following the previous works, we evaluate our method on the standard metrics for anomaly segmentation. 
On the Fishyscapes Lost\&Found and Road Anomaly, we adopt the the area under receiver operating characteristics (AUROC), average precision (AP), and the false positive rate at a true positive rate of 95\% (FPR95). 
On the SMIYC, in addition to reporting per-pixel metrics such as AP and FPR95, we also calculate results for component-level metrics~\cite{chan2021segmentmeifyoucan} such as the component-wise intersection over union (sIoU), positive predictive value (PPV), and mean F1. 

\noindent \textbf{Implementation Details: } To demonstrate the effectiveness of RWPM for anomaly segmentation methods, we selected representative methods from both pixel-based and mask-based networks. For pixel-based methods, we chose PEBAL~\cite{tian2022pixel} and Balanced Energy~\cite{choi2023balanced}. These methods all utilize DeepLabv3+~\cite{chen2018encoder} with WideResnet38 as the backbone and were trained on the Cityscapes dataset and COCO OE images~\cite{tian2022pixel}. For mask-based methods, we selected RbA~\cite{nayal2023rba} and Mask2Anomaly~\cite{rai2023unmasking}. These methods use Mask2Former~\cite{cheng2022masked} with Swin-L~\cite{liu2021swin} as the backbone and were also trained on the Cityscapes and COCO OE images. 
All experiments are conducted on a single NVIDIA RTX A6000 GPU. 
To ensure efficient implementation on a single GPU for all experiments, we adopt embedding map partitioning and limited iteration in all experiments. 
Specifically, for pixel-based methods, we set the partitioning parameter $n = 4$. For mask-based methods, as their embedding map size is smaller, we set $n = 2$. 
Additionally, for the random walk, if not specifically emphasized, we set the number of iterations $T = 20$/$T = 5$ for the Road Anomaly/the other datasets and the transition probability $\alpha = 0.99$ for all datasets. 
The $\tau$ is set as $0.01$. 
It is important to note that our method is used in the inference phase. In other words, RWPM is directly applied to the pre-trained models of other existing anomaly segmentation methods, thus requiring no additional training. 

\subsection{Effectiveness of RWPM}
First, to validate that RWPM can improve the performance of existing anomaly segmentation methods, 
we integrate our approach with four recent representative methods, namely PEBAL~\cite{tian2022pixel}, 
Balanced Energy~\cite{choi2023balanced}, Mask2Anomaly~\cite{rai2023unmasking} and RbA~\cite{nayal2023rba}. 
PEBAL and Balanced Energy employ pixel-based networks as their backbones, whereas Mask2Anomaly and RbA utilize mask-based networks as their backbones. 
We evaluate the results on both the Fishyscapes Lost\&Found validation set and Road Anomaly test set. 
The experimental results are reported in Table~\ref{tab:effect}. 
These results demonstrate that our proposed RWPM consistently and significantly enhances the performance of existing anomaly segmentation methods, across different backbone architectures. 
Additional results and parameters details are shown in Section S.3 of the supplementary material. 

\begin{table}[t]
\tabcolsep=7pt
\caption{Comparison with strong and representative anomaly segmentation baselines across different backbone Architectures (Arch). \textbf{Bold} denotes the better results between using RWPM and not using RWPM with the same anomaly segmentation baselines. \textcolor{red}{$^\dagger$ } indicates that use the calibration in sub-map concatenation as described in Section~\ref{subsec:prw}.}
\centering
\resizebox{0.8\linewidth}{!}{
    \begin{tabular}{l|l|ccc|ccc}
    \toprule
    \multicolumn{2}{l}{Benchmark$\rightarrow$} & \multicolumn{3}{c}{Fishyscapes Lost\&Found} & \multicolumn{3}{c}{Road Anomaly} \\
    \midrule
    Method $\downarrow$ &Arch    & AuROC$\uparrow$          & AP$\uparrow$         & FPR95$\downarrow$         & AuROC$\uparrow$      & AP$\uparrow$      & FPR95$\downarrow$     \\
    \midrule
    PEBAL~\cite{tian2022pixel}  &Pixel-based        &98.96           &58.81       &4.77           &87.63       &45.10    &44.58           \\
    PEBAL + RWPM$\textcolor{red}{^\dagger}$ (Ours) &Pixel-base&\textbf{99.20}  &\textbf{66.85} & \textbf{3.68} &\textbf{89.48} &\textbf{50.29} & \textbf{36.81}          \\
    \midrule
    Balanced Energy~\cite{choi2023balanced}  &Pixel-based   &99.03             & 67.07      & 2.93          & 88.31      &41.48    & 41.46     \\
Balanced Energy + RWPM$\textcolor{red}{^\dagger}$ (Ours) &Pixel-base & \textbf{99.29} & \textbf{72.95} & \textbf{2.38}  & \textbf{90.51} & \textbf{48.26}& \textbf{32.10}           \\
    \midrule
    Mask2Anomaly\tablefootnote[2]{We carefully reproduce the experiments by using the official code.}~\cite{rai2023unmasking}    &Mask-based        & 95.47            & 65.27       &  7.79        & 96.54       & 80.04  &  13.95         \\
    Mask2Anomaly + RWPM (Ours) &Mask-base  &  \textbf{95.48}        &  \textbf{65.38}      & \textbf{7.33}  & \textbf{97.44} & \textbf{80.09} & \textbf{7.45}  \\
    \midrule
    RbA~\cite{nayal2023rba}    &Mask-based                  &  98.62         &   70.81    &   6.30        &  97.99     &  85.42  & 6.92      \\
    RbA + RWPM (Ours) &Mask-base      &  \textbf{98.82}      &   \textbf{71.16}      &  \textbf{6.12}           & \textbf{98.04} &  \textbf{87.34}  & \textbf{5.27} \\
    \bottomrule
    \end{tabular}}
\label{tab:effect}
\end{table}

\begin{table}[bhtp]
\tabcolsep=3pt
\caption{Comparison with state-of-the-art anomaly segmentation methods across different backbone Architectures (Arch). 
\textbf{Bold} and \underline{underline} denote the best and the second best results, respectively. }
\centering
\resizebox{1.0\linewidth}{!}{
    \begin{tabular}{l|l||cc|cc|cc|cc|cc}
    \toprule
   \multicolumn{2}{l}{Benchmark$\rightarrow$} & \multicolumn{2}{c}{Fishyscapes L\&F} & \multicolumn{2}{c}{Road Anomaly} & \multicolumn{2}{c}{Anomaly Track}  & \multicolumn{2}{c}{Obstacle Track} & \multicolumn{2}{c}{Average}\\
    \midrule
    Method $\downarrow$   &Arch  & AP$\uparrow$  & FPR95$\downarrow$  & AP$\uparrow$   & FPR95$\downarrow$  & AP$\uparrow$  & FPR95$\downarrow$  & AP$\uparrow$  & FPR95$\downarrow$   & AP$\uparrow$  & FPR95$\downarrow$    \\
    \midrule
    Synboost~\cite{di2021pixel} (CVPR'21) &Pixel-based             &60.58  &31.02     &41.84 &59.72  &56.44  &61.86  &71.34  &3.15  &57.46  &38.94  \\ 
    SML~\cite{jung2021standardized} (ICCV'21) &Pixel-based         &22.74  &33.49     &25.82 &49.47  &46.8  &39.5  &3.4  &36.8  &24.69  &39.82  \\ 
    Meta-OOD~\cite{chan2021entropy} (ICCV'21) &Pixel-based         &41.31  &37.69     &48.84      &31.77   &85.47  &15.00  &85.07  &0.75  &65.13  &21.30    \\ 
    Learning Embedding~\cite{blum2021fishyscapes}(IJCV'21) &Pixel-based&4.65  &24.36     &-      &-   &37.52  &70.76  &0.82  &46.38  &-  &-    \\
    Void Classifier~\cite{blum2021fishyscapes}(IJCV'21) &Pixel-based&10.29  &22.11     &-      &-   &36.61  &63.49  &10.44  &41.54  &-  &-    \\
    % JSRNet~\cite{vojir2021road} (ICCV'21)   &Pixel-based            &-  &-     &94.4      &     &  &  &  &  &  &    \\
    GMMSeg-DL~\cite{liang2022gmmseg} (NeurIPS’22) &Pixel-based     &43.47  &13.11     &34.42  &47.90 &-  &-  &-  &-  &-  &-  \\ 
    DenseHybrid~\cite{grcic2022densehybrid}  (ECCV'22) &Pixel-based &69.79  &5.09     &31.39  &63.97 &77.96  &\textbf{9.81}  &87.08  &\underline{0.24}  &65.56  &19.78   \\    
    PEBAL~\cite{tian2022pixel} (ECCV'22) &Pixel-based               &58.81  &4.77     &45.10 &44.58  &49.14  &40.82  &4.98  &12.68  &39.58  &25.71     \\
    Balanced Energy~\cite{choi2023balanced} (CVPR'23) &Pixel-based  &67.07  &\underline{2.93}     &41.48 &41.46  &-  &-  &-  &-  &-  &-    \\
    RPL-CoroCL~\cite{liu2023residual} (ICCV'23) &Pixel-based  &70.61  &\textbf{2.52}     &71.61 &17.74  &83.49  &11.68  &85.93  &0.58  &77.91  &8.13    \\
     % \midrule
    Mask2Anomaly~\cite{rai2023unmasking} (ICCV'23) & Mask-based     &69.44  &9.22    &80.04 &13.95   &88.62  &14.57  &\underline{93.10}  &\textbf{0.20}  &82.87  &9.49   \\
    M2F-EAM~\cite{grcic2023advantages} (CVPRW'23) &Mask-based       &52.03  &20.51   &66.67 &13.42   &76.3  &93.9  &66.9  &17.9  &65.48  &36.43     \\
    % Maskomaly~\cite{ackermann2023maskomaly} (BMVC'23) &Mask-based   &-      &-       &70.93 &11.88   &  &  &  &  &  &    \\
    \midrule
    RbA~\cite{nayal2023rba} (ICCV'23) &Mask-based                   &\underline{70.81}  &6.30    &\underline{85.42} &\underline{6.92}    &\underline{90.86}  &11.59  &91.85  &0.46  &\underline{84.73}  &\underline{6.33}    \\
    
     % \midrule

    \textbf{RbA + RWPM (Ours)}       & Mask-based  &\textbf{71.16}      &  6.12   &  \textbf{87.34}  & \textbf{5.27} &\textbf{92.00}  &\underline{10.15}  &\textbf{93.30}  &0.28 &\textbf{86.00}  &\textbf{5.46} \\
    \bottomrule
    \end{tabular}}
\label{tab:result_sota}
\end{table}

Second, to further demonstrate the effectiveness of our RWPM, we compare it with other state-of-the-art (SOTA) methods. 
Specifically, we adopt the combination of RbA with RWPM as our instance approach for the comparison. 
In addition to the Fishyscapes and Road Anomaly datasets, we also evaluate our approach on a more challenging dataset, \ie, SMIYC online test set (Anomaly track and Obstacle track). 
SMIYC is characterized by its high domain shift and diversity of objects. 
Table~\ref{tab:result_sota} presents the results. 
The results indicate that our method can consistently achieve high performance across various datasets, while the previous SOTA methods fail to do so. 
We also compare all approaches by computing the average results of AP and FPR95. The results shows that our approach outperforms the previous methods and achieve the best results. 

Furthermore, in addition to pixel-level metrics, we also utilize component-level metrics~\cite{chan2021segmentmeifyoucan} to evaluate performance on the SMIYC online test set. 
Compared to pixel-level metrics, component-level metrics provide a better evaluation of the detection results for all anomalous objects in the scene. 
We report the results in Table~\ref{tab:result_smiyc}. 
The experimental results show that RWPM significantly improves the performance of RbA in component-level metrics, surpassing other existing SOTA methods by a large margin. 
This indicates that RWPM can assist models in better detecting anomalous objects and achieving more precise localization of these anomalies~\cite{chan2021segmentmeifyoucan}.

\begin{table}[thbp]
\tabcolsep=10pt
\caption{Component level evaluation: Comparison with state-of-the-art anomaly segmentation methods on the SMIYC online test set. 
\textbf{Bold} denote the best results.  
}
\centering
\resizebox{1.0\linewidth}{!}{
    \begin{tabular}{l|l||ccc|ccc}
    \toprule
   \multicolumn{2}{l}{Benchmark$\rightarrow$} & \multicolumn{3}{c}{Anomaly Track} & \multicolumn{3}{c}{Obstacle Track} \\
    \midrule
     Method $\downarrow$   &Arch &sIoU $\uparrow$ &PPV$\uparrow$ &mean F1$\uparrow$    &sIoU $\uparrow$ &PPV$\uparrow$ &mean F1$\uparrow$ \\
    \midrule
    Synboost~\cite{di2021pixel} (CVPR'21) &Pixel-based                &34.68 &17.81 &9.99 &44.28 &41.75 &37.57   \\ 
    SML~\cite{jung2021standardized} (ICCV'21) &Pixel-based            &26.00 &24.70 &12.20 &5.10 &13.30 &3.00   \\ 
    Meta-OOD~\cite{chan2021entropy} (ICCV'21) &Pixel-based            &49.31 &39.51 &28.72 &47.87 &63.64 &48.51   \\ 
    Void Classifier~\cite{blum2021fishyscapes}(IJCV'21) &Pixel-based  &21.14 &22.13 &6.49 &6.34 &20.27 &5.41   \\ 
    Learning Embedding~\cite{blum2021fishyscapes}(IJCV'21) &Pixel-based  &33.86 &20.54 &7.90 &35.64 &2.87 &2.31   \\ 
    JSRNet~\cite{vojir2021road} (ICCV'21)   &Pixel-based              &20.20 &29.27 &13.66 &18.55 &24.46 &11.02   \\ 
    % GMMSeg-DL~\cite{liang2022gmmseg} (NeurIPS’22) &Pixel-based        & & & & & &   \\ 
    DenseHybrid~\cite{grcic2022densehybrid}  (ECCV'22) &Pixel-based   &54.17 &24.13 &31.08 &45.74 &50.10 &50.72   \\    
    PEBAL~\cite{tian2022pixel} (ECCV'22) &Pixel-based                 &38.88 &27.20 &14.48 &29.91 &7.55 &5.54   \\ 
    % Balanced Energy~\cite{choi2023balanced} (CVPR'23) &Pixel-based    & & & & & &   \\ 
    RPL-CoroCL~\cite{liu2023residual} (ICCV'23) &Pixel-based          &49.76 &29.96 &30.16 &52.61 &56.65 &56.69   \\ 
    Mask2Fromer~\cite{rai2023unmasking} (ICCV'23) &Mask-based         &25.20 &18.20 &15.30 &5.00 &21.90 &4.80   \\ 
    % Mask2Anomaly~\cite{rai2023unmasking} (ICCV'23) & Mask-based       &60.40 &45.70 &48.60 &55.35 &70.30 &69.80   \\   
    Maskomaly~\cite{ackermann2023maskomaly} (BMVC'23) &Mask-based     &55.4  &51.6  &50.0  &- &- &-   \\   
    \midrule
    RbA~\cite{nayal2023rba} (ICCV'23) &Mask-based                     &55.69  &52.14  &46.80  &58.36  &58.78  &60.85   \\
    \textbf{RbA + RWPM (Ours)}  &Mask-based                      &\textbf{57.00} &\textbf{61.25} &\textbf{58.44} &\textbf{58.89} &\textbf{72.51} &\textbf{69.85}   \\

    \bottomrule
    \end{tabular}}
\label{tab:result_smiyc}
\end{table}

\subsection{Ablation Study} 
\label{sec:ablation}
If not specifically emphasized, all the results reported in this section are conducted on a RTX A6000 GPU. 

\noindent \textbf{The effect of the partitioning parameter $n$ and calibration:} We first test the effect of the embedding map partitioning parameter $n$ and the calibration, and we report the results in Table~\ref{tab:partition}. 
FPS denotes the frames-per-second results. 
Row $n=1$ displays the RWPM results without utilizing the embedding map partitioning strategy, indicating significant computational and memory overhead. 
However, the partitioning strategy ($n=2,4,8$)  significantly improves the running speed and reduces memory consumption while maintaining effectiveness. Additionally, as $n$ increases, the running efficiency also improves, albeit with a slight decrease in performance. However, when $n>2$, the calibration can enhance the performance, particularly for the FPR95 metrics.

\begin{table}[htbp]
\tabcolsep=10pt
\caption{The effect of the partitioning parameter $n$ and calibration. 
We set $T = 20$ for all experiments. 
\textcolor{red}{$^\dagger$ } indicates that use the calibration in sub-map concatenation.}
\centering
\resizebox{0.6\linewidth}{!}{
    \begin{tabular}{lcccc}
    \toprule
    \multicolumn{1}{l}{Benchmark$\rightarrow$} & \multicolumn{4}{c}{Road Anomaly}  \\
    \midrule
     & \textbf{AP$\uparrow$}    & \textbf{FPR95$\downarrow$}  & \textbf{FPS$\uparrow$}    & \textbf{GPU Memory Usage$\downarrow$}   \\
    \midrule
    RbA w/o RWPM  &85.42 &6.92 &11.12 & 3.49GiB   \\
    \midrule
    n=1 (two GPUs)   &87.90 &5.17 &0.32 & 74.96GiB   \\
    \midrule
    n=2  &87.34 &5.27 &2.04 &7.24GiB   \\
    n=2$\textcolor{red}{^\dagger}$  &87.18 &5.25 &2.04 &7.24GiB   \\
    \midrule
    n=4  &87.05 &6.31 &4.35 &3.49GiB   \\
    n=4$\textcolor{red}{^\dagger}$  &87.17 &5.32 &4.35 &3.49GiB   \\
    \midrule
    n=8  &86.67 &6.42 &6.26 &3.55GiB   \\
    n=8$\textcolor{red}{^\dagger}$  &86.91 &5.41 &6.26 &3.55GiB   \\
    
    \bottomrule
    \end{tabular}}
\label{tab:partition}
\end{table}

\noindent \textbf{The effect of iteration number $T$:}
Table~\ref{tab:iteration_number} shows that, compared to using Eq.~\ref{eq:random_walk_inf} ($T=\infty$), utilizing the limited iteration strategy ($T=5\sim100$) can lead to higher efficiency, resulting in even better performance for RWPM. 
Additionally, as the number of iterations increases, performance tends to improve, but running efficiency (FPS) decreases accordingly. 

\begin{table}[htbp]
\tabcolsep=10pt
\caption{The effect of iteration number $T$. 
We set $n=2$ for all experiments. }
\centering
\resizebox{0.6\linewidth}{!}{
    \begin{tabular}{lcccc}
    \toprule
    \multicolumn{1}{l}{Benchmark$\rightarrow$} & \multicolumn{4}{c}{Road Anomaly}  \\
    \midrule
     & \textbf{AP$\uparrow$}    & \textbf{FPR95$\downarrow$}  & \textbf{FPS$\uparrow$}    & \textbf{GPU Memory Usage$\downarrow$}   \\
    \midrule
    RbA w/o RWPM  &85.42 &6.92 &11.12 & 3.49GiB   \\
    \midrule
    T=$\infty$   &87.51 &5.24 &0.29 &8.88GiB    \\
    \midrule
    T=5   &86.89 &5.98 &3.70 & 7.24GiB   \\
    T=10   &87.08 &5.77 &2.78 & 7.24GiB   \\
    T=20   &87.34 &5.27 &2.04 & 7.24GiB   \\
    T=50   &87.36 &5.08 &0.92 &  7.24GiB  \\
    T=100  &87.54 &5.02 &0.50 &  7.24GiB  \\   
    \bottomrule
    \end{tabular}}
\label{tab:iteration_number}
\end{table}

\noindent \textbf{Different strategies for reducing the number of pixel embedding:} 
We also attempt other strategies to reduce the number of pixel embedding for RWPM, such as resizing the embedding map or use SLIC~\cite{achanta2012slic} algorithm to extract superpixels.  
Table~\ref{tab:reduce_pixel}(Left) presents the results. 
The resize strategy achieves good performance but requires preserving a larger map size. Therefore, it fails to substantially reduce the number of pixel embeddings, resulting in significant GPU memory usage. 
On the other hand, the superpixel strategy requires a considerable amount of runtime for the superpixel extraction. 
In comparison, our proposed partitioning strategy can achieve good results with higher efficiency and lower resource consumption. This makes it more suitable for applications on practical platforms.

\noindent \textbf{In-distribution segmentation performance on Cityscapes validation set:} 
We report the results on the Cityscapes validation set in Table~\ref{tab:reduce_pixel}(Right). 
All methods use Mask2Former~\cite{cheng2022masked} as their backbones. 
We observe that most SOTA methods affect the in-distribution segmentation performance of Mask2Former, such as RbA (drop from $82.25$ to $81.93$). 
However, RWPM can alleviate this problem.  
Additionally, even for the SOTA segmentation method (Mask2Former), RWPM can further enhance its in-distribution segmentation performance.

\begin{table}[htbp]
    \caption{\textbf{Ablation Tables:} \textbf{(Left)} shows the results of different strategies for reducing the number of pixel embedding. We find that partitioning is the best. \textbf{(Right)} shows the segmentation performance on the Cityscapes validation set. RWPM can also improve the segmentation performance of in-distribution.}
    \begin{subtable}{0.5\linewidth}
    \tabcolsep=10pt
      \centering
        % \caption{}
        \resizebox{1.0\linewidth}{!}{
        \begin{tabular}{l|cccc}
        \toprule
        \multicolumn{1}{l}{Benchmark$\rightarrow$} & \multicolumn{4}{c}{Road Anomaly}  \\
        \midrule
         Strategy$\downarrow$& \textbf{AP$\uparrow$}    & \textbf{FPR95$\downarrow$}  & \textbf{FPS$\uparrow$}  & Memory \\
        \midrule
        Resize (0.9 full size)  &87.80 &5.39 &1.41 &23.82GiB  \\
        Resize (0.8 full size)  &87.24 &5.44 &2.15 &15.98GiB  \\
        \midrule
        Superpixel (30000 pixels)  &87.51 &5.50 &0.19 &  11.46GiB\\
        Superpixel (20000 pixels)  &87.03 &5.98 &0.32 & 6.31GiB \\
        \midrule
        Partitioning ($n=2$)  &87.34 &5.27 &2.04 &7.24GiB  \\  
        \bottomrule
        \end{tabular}}
    \end{subtable}%
    \begin{subtable}{.5\linewidth}
    \tabcolsep=13pt
      \centering
        % \caption{}
        \resizebox{0.68\linewidth}{!}{
        \begin{tabular}{l|c}
        \toprule
        \multicolumn{1}{l}{Benchmark$\rightarrow$} & \multicolumn{1}{c}{Cityscapes} \\
        \midrule
         Method$\downarrow$& \textbf{mIoU$\uparrow$}  \\
        \midrule
        PEBAL-Mask~\cite{nayal2023rba}  &75.32   \\
        DenseHybrid-Mask~\cite{nayal2023rba}  &80.27   \\
        \midrule
        RbA~\cite{nayal2023rba}  &81.93  \\
        RbA+RWPM(Ours)  &\textbf{82.16}   \\
        \midrule
        Mask2Former~\cite{cheng2022masked}  &82.25   \\
        Mask2Former+RWPM(Ours) &\textbf{82.43}   \\
        \bottomrule
        \end{tabular}}
    \end{subtable} 
\label{tab:reduce_pixel}
\end{table}

\subsection{Visualization of Pixel Embedding Distribution}
\begin{figure}[htbp]
    \centering
    \begin{tabular}{ccc}
        \includegraphics[width=0.28\linewidth]{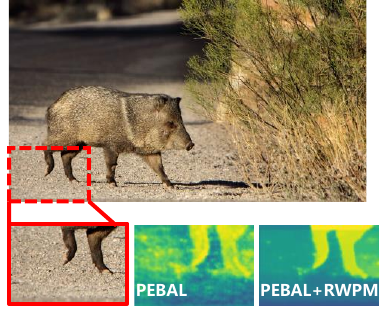} &
	\includegraphics[width=0.36\linewidth]{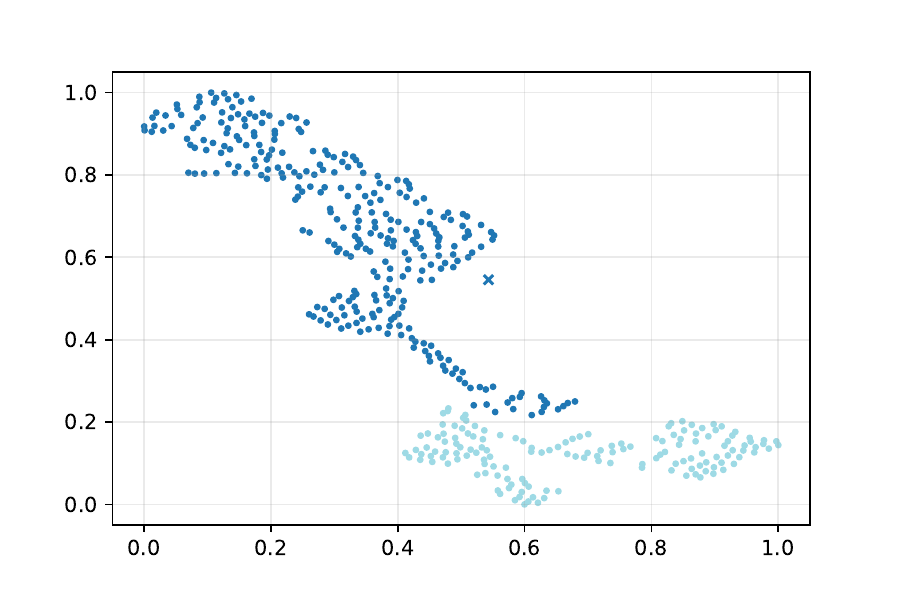} &
	\includegraphics[width=0.36\linewidth]{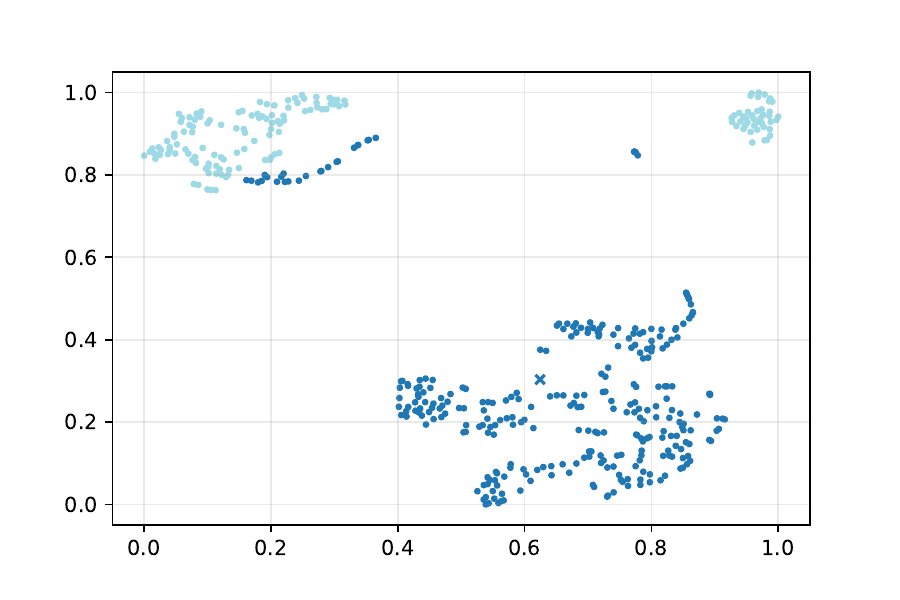} \\
    (a) Sample Image & (b) Distribution w/o RWPM & (c) Distribution with RWPM
    \end{tabular}
    \caption{\textbf{The visualization of pixel embedding distribution.} All embeddings are extracted from PEBAL model. We use cosine distance as the distance metric. We observed that RWPM can optimize the distribution of pixel embeddings in the space.}
    \label{fig:visualization}
\end{figure}

To investigate the effect of RWPM in the embedding space, we visualized the distribution of pixel embeddings by using t-SNE. 
We selected an image from the Road Anomaly dataset (as shown in Fig~\ref{fig:visualization}(a)). 
Due to the gravel road surface in this image, there is a significant domain gap between the road in this image and the roads in the training set, resulting in many false positives on the road (as shown in the PEBAL results in Fig~\ref{fig:visualization}(a)). 
Then, we randomly selected 300 inlier pixels and 150 outlier pixels from the red dashed box region of this image for visualization. 
Fig~\ref{fig:visualization}(b) and Fig~\ref{fig:visualization}(c) respectively show the distributions of these pixels in the space without and with applying RWPM. 
In Fig~\ref{fig:visualization}(b) and Fig~\ref{fig:visualization}(c), deep blue circles represent inlier pixels (road), light blue circles represent outlier pixels (boar), and X symbols represent the inlier class prototype (the vector of network weights of classifier). 
From Fig~\ref{fig:visualization}(b), we observe that without using RWPM, due to the influence of data diversity, the distribution of inliers and outliers is distorted. 
Consequently, it is difficult for an anomaly scoring function to distinguish between inlier and outlier pixels based on their logits. 
However, from Fig~\ref{fig:visualization}(c), we find that RWPM can significantly alleviate the problem. 
Specifically, the pixel embeddings of the same category form more compact clusters, enabling the anomaly scoring function to clearly discriminate between inliers and outliers based on the logit. 
This is consistent with the description in the toy sample in Fig~\ref{fig:manifolds1}. 
Note that, in the top left corner of Fig~\ref{fig:visualization}(c), some inlier points become very close to outlier points after applying RWPM. 
This is because these points are located in the boundaries of anomalous objects, making it very difficult to discriminate them.

\subsection{Qualitative Result}

\begin{figure}[htbp]
  \centering
  \includegraphics[width=0.95\textwidth]{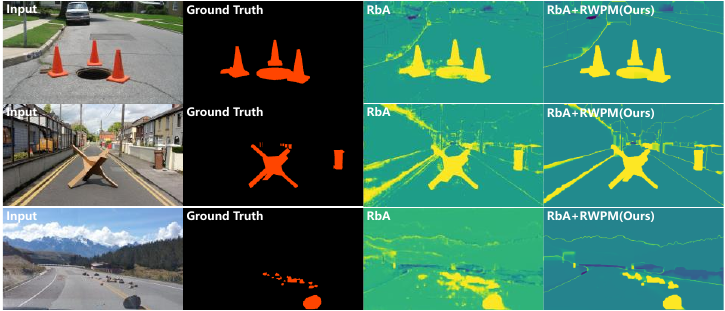}
  \caption{\textbf{Qualitative results: }The first column displays the input image, while the second column shows its corresponding ground truth. The third column displays the anomaly score results generated by RbA, and the fourth column presents the anomaly score results after applying RWPM to RbA. Yellow indicates high anomaly scores.}
  \label{fig:qualitative}
\end{figure}

Finally, to visually demonstrate the effectiveness of RWPM, we present qualitative results. Specifically, we compare the results of RbA on the Road Anomaly dataset without and with RWPM. 
The qualitative results are illustrated in Fig~\ref{fig:qualitative}.
Fig~\ref{fig:qualitative} shows that RWPM significantly improves the performance of RbA, reducing both the false positive rate and false negative rate. 
For example, in the first row, RbA generates some false positive results on the road and some false negative results on the manhole region (third column of the first row). 
However, after applying RWPM, these false positive/negative results are effectively eliminated (fourth column of the first row). 
Additionally, RWPM assists the RbA model to accurately detect and localize each anomalous object at the component level, such as more precise edge detection. This observation aligns with the results of the component-level metrics presented in Table~\ref{tab:result_smiyc}.

\section{Conclusions and Discussions}
In this paper, we propose a simple yet effective method, called Random Walk on Pixel Manifolds (RWPM), to improve the performance of existing anomaly segmentation methods. 
RWPM is a post-processing method that can be directly integrated into existing anomaly detection frameworks without requiring additional training or modifications to the network structure. 
RWPM mitigates the impact of data diversity on manifold structure by diffusing and updating pixel embeddings on pixel manifolds using random walk. 
This process enables anomaly scoring functions to obtain more accurate estimations of anomaly scores. 
Additionally, we propose Partial Random Walk strategy, which significantly improve the efficiency of RWPM. 

While RWPM achieves significant results in most cases, it is limited when the pre-trained model itself exhibits significant inaccuracies. 
To address this limitation, we plan to enhance the model's generalization ability during training to accurately detect a broader range of diverse and unknown anomalies. 

\section*{Acknowledgements}
This work was supported by SenseTime Japan.

\bibliographystyle{splncs04}
\bibliography{main}

\begin{thebibliography}{10}
\providecommand{\url}[1]{\texttt{#1}}
\providecommand{\urlprefix}{URL }
\providecommand{\doi}[1]{https://doi.org/#1}

\bibitem{achanta2012slic}
Achanta, R., Shaji, A., Smith, K., Lucchi, A., Fua, P., S{\"u}sstrunk, S.: Slic superpixels compared to state-of-the-art superpixel methods. IEEE transactions on pattern analysis and machine intelligence  \textbf{34}(11),  2274--2282 (2012)

\bibitem{ackermann2023maskomaly}
Ackermann, J., Sakaridis, C., Yu, F.: Maskomaly: Zero-shot mask anomaly segmentation. arXiv preprint arXiv:2305.16972  (2023)

\bibitem{bai2017regularized}
Bai, S., Bai, X., Tian, Q., Latecki, L.J.: Regularized diffusion process for visual retrieval. In: Proceedings of the AAAI conference on artificial intelligence. vol.~31 (2017). \doi{10.1609/aaai.v31i1.11216}

\bibitem{bai2017ensemble}
Bai, S., Zhou, Z., Wang, J., Bai, X., Jan~Latecki, L., Tian, Q.: Ensemble diffusion for retrieval. In: Proceedings of the IEEE International conference on computer vision. pp. 774--783 (2017)

\bibitem{blum2021fishyscapes}
Blum, H., Sarlin, P.E., Nieto, J., Siegwart, R., Cadena, C.: The fishyscapes benchmark: Measuring blind spots in semantic segmentation. International Journal of Computer Vision  \textbf{129},  3119--3135 (2021). \doi{10.1007/s11263-021-01511-6}

\bibitem{chan2021segmentmeifyoucan}
Chan, R., Lis, K., Uhlemeyer, S., Blum, H., Honari, S., Siegwart, R., Fua, P., Salzmann, M., Rottmann, M.: Segmentmeifyoucan: A benchmark for anomaly segmentation. arXiv preprint arXiv:2104.14812  (2021)

\bibitem{chan2021entropy}
Chan, R., Rottmann, M., Gottschalk, H.: Entropy maximization and meta classification for out-of-distribution detection in semantic segmentation. In: Proceedings of the ieee/cvf international conference on computer vision. pp. 5128--5137 (2021)

\bibitem{chen2017deeplab}
Chen, L.C., Papandreou, G., Kokkinos, I., Murphy, K., Yuille, A.L.: Deeplab: Semantic image segmentation with deep convolutional nets, atrous convolution, and fully connected crfs. IEEE transactions on pattern analysis and machine intelligence  \textbf{40}(4),  834--848 (2017)

\bibitem{chen2018encoder}
Chen, L.C., Zhu, Y., Papandreou, G., Schroff, F., Adam, H.: Encoder-decoder with atrous separable convolution for semantic image segmentation. In: Proceedings of the European conference on computer vision (ECCV). pp. 801--818 (2018). \doi{10.1007/978-3-030-01234-2_49}

\bibitem{cheng2022masked}
Cheng, B., Misra, I., Schwing, A.G., Kirillov, A., Girdhar, R.: Masked-attention mask transformer for universal image segmentation. In: Proceedings of the IEEE/CVF conference on computer vision and pattern recognition. pp. 1290--1299 (2022). \doi{10.1109/CVPR52688.2022.00135}

\bibitem{cheng2021per}
Cheng, B., Schwing, A., Kirillov, A.: Per-pixel classification is not all you need for semantic segmentation. Advances in Neural Information Processing Systems  \textbf{34},  17864--17875 (2021)

\bibitem{choi2023balanced}
Choi, H., Jeong, H., Choi, J.Y.: Balanced energy regularization loss for out-of-distribution detection. In: Proceedings of the IEEE/CVF Conference on Computer Vision and Pattern Recognition. pp. 15691--15700 (2023)

\bibitem{cordts2016cityscapes}
Cordts, M., Omran, M., Ramos, S., Rehfeld, T., Enzweiler, M., Benenson, R., Franke, U., Roth, S., Schiele, B.: The cityscapes dataset for semantic urban scene understanding. In: Proceedings of the IEEE conference on computer vision and pattern recognition. pp. 3213--3223 (2016). \doi{10.1109/CVPR.2016.350}

\bibitem{creusot2015real}
Creusot, C., Munawar, A.: Real-time small obstacle detection on highways using compressive rbm road reconstruction. In: 2015 IEEE Intelligent Vehicles Symposium (IV). pp. 162--167. IEEE (2015)

\bibitem{di2021pixel}
Di~Biase, G., Blum, H., Siegwart, R., Cadena, C.: Pixel-wise anomaly detection in complex driving scenes. In: Proceedings of the IEEE/CVF conference on computer vision and pattern recognition. pp. 16918--16927 (2021). \doi{10.1109/CVPR46437.2021.01664}

\bibitem{gal2016dropout}
Gal, Y., Ghahramani, Z.: Dropout as a bayesian approximation: Representing model uncertainty in deep learning. In: international conference on machine learning. pp. 1050--1059. PMLR (2016)

\bibitem{grady2006random}
Grady, L.: Random walks for image segmentation. IEEE transactions on pattern analysis and machine intelligence  \textbf{28}(11),  1768--1783 (2006)

\bibitem{grcic2022densehybrid}
Grci{\'c}, M., Bevandi{\'c}, P., {\v{S}}egvi{\'c}, S.: Densehybrid: Hybrid anomaly detection for dense open-set recognition. In: European Conference on Computer Vision. pp. 500--517. Springer (2022)

\bibitem{grcic2023advantages}
Grci{\'c}, M., {\v{S}}ari{\'c}, J., {\v{S}}egvi{\'c}, S.: On advantages of mask-level recognition for outlier-aware segmentation. In: Proceedings of the IEEE/CVF Conference on Computer Vision and Pattern Recognition. pp. 2936--2946 (2023)

\bibitem{hendrycks2019scaling}
Hendrycks, D., Basart, S., Mazeika, M., Zou, A., Kwon, J., Mostajabi, M., Steinhardt, J., Song, D.: Scaling out-of-distribution detection for real-world settings. arXiv preprint arXiv:1911.11132  (2019)

\bibitem{hendrycks2017a}
Hendrycks, D., Gimpel, K.: A baseline for detecting misclassified and out-of-distribution examples in neural networks. In: International Conference on Learning Representations (2017), \url{https://openreview.net/forum?id=Hkg4TI9xl}

\bibitem{hendrycks2018deep}
Hendrycks, D., Mazeika, M., Dietterich, T.: Deep anomaly detection with outlier exposure. arXiv preprint arXiv:1812.04606  (2018)

\bibitem{huang2019difference}
Huang, Z., Li, S.: From difference to similarity: A manifold ranking-based hyperspectral anomaly detection framework. IEEE Transactions on Geoscience and Remote Sensing  \textbf{57}(10),  8118--8130 (2019)

\bibitem{iscen2018mining}
Iscen, A., Tolias, G., Avrithis, Y., Chum, O.: Mining on manifolds: Metric learning without labels. In: Proceedings of the IEEE Conference on Computer Vision and Pattern Recognition. pp. 7642--7651 (2018). \doi{10.1109/CVPR.2018.00797}

\bibitem{iscen2017efficient}
Iscen, A., Tolias, G., Avrithis, Y., Furon, T., Chum, O.: Efficient diffusion on region manifolds: Recovering small objects with compact cnn representations. In: Proceedings of the IEEE conference on computer vision and pattern recognition. pp. 2077--2086 (2017). \doi{10.1109/CVPR.2017.105}

\bibitem{jung2021standardized}
Jung, S., Lee, J., Gwak, D., Choi, S., Choo, J.: Standardized max logits: A simple yet effective approach for identifying unexpected road obstacles in urban-scene segmentation. In: Proceedings of the IEEE/CVF International Conference on Computer Vision. pp. 15425--15434 (2021)

\bibitem{lakshminarayanan2017simple}
Lakshminarayanan, B., Pritzel, A., Blundell, C.: Simple and scalable predictive uncertainty estimation using deep ensembles. Advances in neural information processing systems  \textbf{30} (2017)

\bibitem{liang2022gmmseg}
Liang, C., Wang, W., Miao, J., Yang, Y.: Gmmseg: Gaussian mixture based generative semantic segmentation models. Advances in Neural Information Processing Systems  \textbf{35},  31360--31375 (2022)

\bibitem{liang2018enhancing}
Liang, S., Li, Y., Srikant, R.: Enhancing the reliability of out-of-distribution image detection in neural networks. In: International Conference on Learning Representations (2018), \url{https://openreview.net/forum?id=H1VGkIxRZ}

\bibitem{lin2014microsoft}
Lin, T.Y., Maire, M., Belongie, S., Hays, J., Perona, P., Ramanan, D., Doll{\'a}r, P., Zitnick, C.L.: Microsoft coco: Common objects in context. In: Computer Vision--ECCV 2014: 13th European Conference, Zurich, Switzerland, September 6-12, 2014, Proceedings, Part V 13. pp. 740--755. Springer (2014). \doi{10.1007/978-3-319-10602-1_48}

\bibitem{lis2019detecting}
Lis, K., Nakka, K., Fua, P., Salzmann, M.: Detecting the unexpected via image resynthesis. In: Proceedings of the IEEE/CVF International Conference on Computer Vision. pp. 2152--2161 (2019). \doi{10.1109/ICCV.2019.00224}

\bibitem{liu2020energy}
Liu, W., Wang, X., Owens, J., Li, Y.: Energy-based out-of-distribution detection. Advances in neural information processing systems  \textbf{33},  21464--21475 (2020)

\bibitem{liu2023residual}
Liu, Y., Ding, C., Tian, Y., Pang, G., Belagiannis, V., Reid, I., Carneiro, G.: Residual pattern learning for pixel-wise out-of-distribution detection in semantic segmentation. In: Proceedings of the IEEE/CVF International Conference on Computer Vision. pp. 1151--1161 (2023)

\bibitem{liu2021swin}
Liu, Z., Lin, Y., Cao, Y., Hu, H., Wei, Y., Zhang, Z., Lin, S., Guo, B.: Swin transformer: Hierarchical vision transformer using shifted windows. In: Proceedings of the IEEE/CVF international conference on computer vision. pp. 10012--10022 (2021)

\bibitem{nayal2023rba}
Nayal, N., Yavuz, M., Henriques, J.F., G{\"u}ney, F.: Rba: Segmenting unknown regions rejected by all. In: Proceedings of the IEEE/CVF International Conference on Computer Vision. pp. 711--722 (2023)

\bibitem{ohgushi2020road}
Ohgushi, T., Horiguchi, K., Yamanaka, M.: Road obstacle detection method based on an autoencoder with semantic segmentation. In: Proceedings of the Asian Conference on Computer Vision (2020)

\bibitem{rai2023unmasking}
Rai, S.N., Cermelli, F., Fontanel, D., Masone, C., Caputo, B.: Unmasking anomalies in road-scene segmentation. In: Proceedings of the IEEE/CVF International Conference on Computer Vision. pp. 4037--4046 (2023)

\bibitem{tian2022pixel}
Tian, Y., Liu, Y., Pang, G., Liu, F., Chen, Y., Carneiro, G.: Pixel-wise energy-biased abstention learning for anomaly segmentation on complex urban driving scenes. In: European Conference on Computer Vision. pp. 246--263. Springer (2022). \doi{10.1007/978-3-031-19842-7_15}

\bibitem{vojir2021road}
Vojir, T., {\v{S}}ipka, T., Aljundi, R., Chumerin, N., Reino, D.O., Matas, J.: Road anomaly detection by partial image reconstruction with segmentation coupling. In: Proceedings of the IEEE/CVF International Conference on Computer Vision. pp. 15651--15660 (2021)

\bibitem{xie2021segformer}
Xie, E., Wang, W., Yu, Z., Anandkumar, A., Alvarez, J.M., Luo, P.: Segformer: Simple and efficient design for semantic segmentation with transformers. Advances in Neural Information Processing Systems  \textbf{34},  12077--12090 (2021)

\bibitem{yang2019efficient}
Yang, F., Hinami, R., Matsui, Y., Ly, S., Satoh, S.: Efficient image retrieval via decoupling diffusion into online and offline processing. In: Proceedings of the AAAI conference on artificial intelligence. vol.~33, pp. 9087--9094 (2019). \doi{10.1609/aaai.v33i01.33019087}

\bibitem{yang2020mining}
Yang, F., Wang, Z., Xiao, J., Satoh, S.: Mining on heterogeneous manifolds for zero-shot cross-modal image retrieval. In: Proceedings of the AAAI Conference on Artificial Intelligence. vol.~34, pp. 12589--12596 (2020)

\bibitem{zhou2003learning}
Zhou, D., Bousquet, O., Lal, T., Weston, J., Sch{\"o}lkopf, B.: Learning with local and global consistency. Advances in neural information processing systems  \textbf{16} (2003)

\bibitem{zhou2003ranking}
Zhou, D., Weston, J., Gretton, A., Bousquet, O., Sch{\"o}lkopf, B.: Ranking on data manifolds. Advances in neural information processing systems  \textbf{16} (2003)

\end{thebibliography}

\title{Supplementary Material for ``Random Walk on Pixel Manifolds for Anomaly Segmentation of Complex Driving Scenes''} 

\titlerunning{Supplementary Material for RWPM}

\author{Zelong Zeng\inst{1}\orcidlink{0000-0003-3740-2743} \and
Kaname Tomite \inst{1}}

\authorrunning{Z.~Zeng et al.}

\institute{SenseTime Japan, Tokyo, Japan\\
\email{\{zengzelong,tomite\}@sensetime.jp}}

\maketitle

\begin{abstract}
    The following items are included in the supplementary material. 
    \begin{itemize}
    \item[-] The Architecture of Pixel-based and Mask-based Networks. 
    \item[-] Anomaly Scoring Function Details. 
    \item[-] Additional Results and Implementation Details. 
    \item[-] Different Methods for Constructing Locally Constrained Graph. 
    \item[-] Analysis for RWPM. 
    \item[-] Some Reviewers' Comments and Our Explanations. 
    \end{itemize}

\end{abstract}

\section{The Architecture of Pixel-based and Mask-based Networks}

In this section, we introduce the structures of pixel-based and mask-based segmentation models, respectively. 
Additionally, we also illustrate the position of the pixel embedding map used by RWPM in different network architectures.
In the main paper, we utilize  the Deeplabv3+~\cite{chen2018encoder} with WideResNet38 as the pixel-based segmentation model, 
and the Mask2Former~\cite{cheng2022masked} with Swin-L as the mask-based segmentation model. 

\begin{figure}[htbp]
  \centering
  \includegraphics[width=0.95\textwidth]{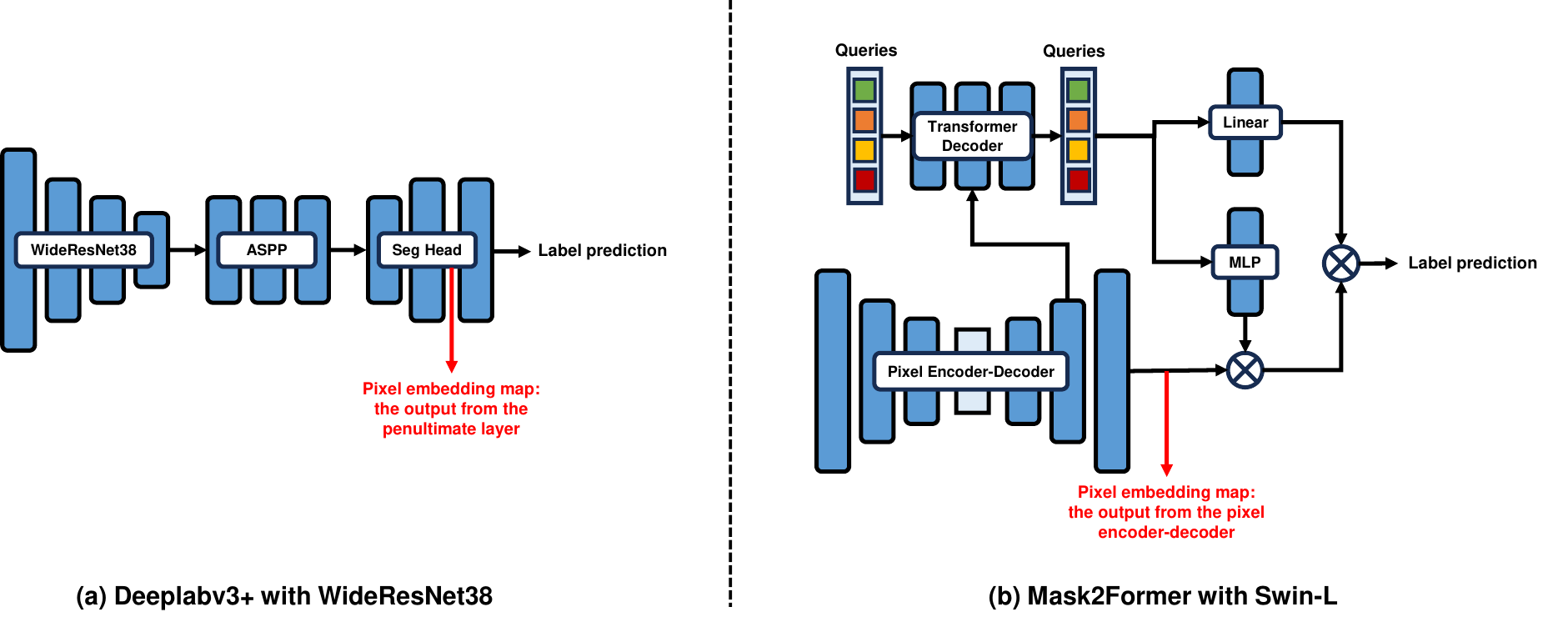}
  \caption{\textbf{The architecture of pixel-based and mask-based networks.} (a) depicts the architecture of the pixel-based network used in the main paper, namely Deeplabv3+ with WideResNet38. (b) illustrates the architecture of the mask-based network used in the main paper, which is Mask2Former with Swin-L. The red arrows indicate the position of the pixel embeddings map used by RWPM in different network architectures. }
  \label{fig:architecture}
\end{figure}

Fig~\ref{fig:architecture} illustrates their architecture. 
Specifically, Fig~\ref{fig:architecture}(a) shows the architecture of the Deeplabv3+ with WideResNet38, which consists of three parts: a WideResNet38 backbone, a Atrous Spatial Pyramid Pooling (ASPP) layers and a Segmentation Head (Seg Head). 
During inference, the input image will be fed into the backbone first to produce a feature map. Then, this feature map will go through the ASPP layers to output a feature map with multiscale information. Finally, such feature map will be classified by the Seg Head to produce the dense predictions. The last layer of Seg Head is a classifier. 
In the main paper, we use the output from the penultimate layer of the Seg Head as the pixel embedding map for RWPM. 

Fig~\ref{fig:architecture}(b) depicts the architecture of the Mask2Former with Swin-L, consisting of a Pixel Encoder-Decoder, a Transformer Decoder, a Linear Classifier and a Multilayer Perceptron (MLP). 
During inference, the input image is first fed into the Pixel Encoder-Decoder to generate a feature map. 
Simultaneously, the intermediate upsampled features from the Pixel Encoder-Decoder are fed to the Transformer Decoder. 
Next, $N$ learnable queries go through the Transformer Decoder with the intermediate upsampled features to produce $N$ refined queries. 
These $N$ refined queries are then fed into the Linear Classifier and MLP. 
The Linear Classifier is the classifier of queries, generating the class predictions of each refined query. 
Meanwhile, the MLP adjusts the dimension of each refined queries to match the size of the feature map from the Pixel Encoder-Decoder. 
After going through the MLP, the $N$ refined queries can be seen as $N$ binary classifiers, which are multiplied with the feature map from the Pixel Encoder-Decoder to produce the mask predictions. 
Finally, by combining the mask predictions with the corresponding class predictions, the model outputs the classification predictions for each pixel of the input image. 
In summary, each refined query generates a mask on the image, where the prediction scores on the mask represent the binary classification results of corresponding pixels, describing the confidence level of the refined query for that pixel. Then, the Linear Classifier categorizes each refined query, obtaining classification predictions for the corresponding refined queries. 
The classification prediction for each pixel is obtained by the weighted sum of the classification predictions of all refined queries. The weight is the mask prediction score generated by each refined query for that pixel. 
In other words, the higher the confidence level of a refined query for a pixel, the more similar the classification result of that pixel is to the classification result of the refined query. 
In the main paper, we use the output from the Pixel Encoder-Decoder as the pixel embeddings map for RWPM. 

\section{Anomaly Scoring Function Details}

In our main paper, we integrate our RWPM with four recent representative anomaly segmentation methods, namely PEBAL~\cite{tian2022pixel}, Balanced Energy~\cite{choi2023balanced}, RbA~\cite{nayal2023rba} and Mask2Anomaly~\cite{rai2023unmasking}. 
In this section, we provide a detailed introduction to the anomaly scoring functions of these methods.

\subsection{PEBAL and Balanced Energy}
PEBAL and Balanced Energy utilize Deeplabv3+ (see Fig~\ref{fig:architecture}(a)), a pixel-based network, as their backbones. 
Both methods utilize the EBM function~\cite{liu2020energy} as their anomaly scoring function. 
Let's define $\boldsymbol{l}_{h,w}\left(k\right)$ as the logit of a pixel $\boldsymbol{x}_{h,w}$ for class $k$. 
$\boldsymbol{l}_{h,w}\left(k\right)$ is calculated from the inner product of the pixel embedding and the vector of network weights for class $k$ of the classifier. 
The anomaly scoring function at the pixel $\boldsymbol{x}_{h,w}$ is defined as: 

\begin{equation}
    E(\boldsymbol{x})_{h,w}=-\log \sum_{k\in\left\{1\cdots K\right\}} \exp\left(\boldsymbol{l}_{h,w}\left(k\right)\right)
    ,
  \label{eq:pebal}
\end{equation}
where $K$ is the number of inlier class. 
Obviously, when the anomaly score $E(\boldsymbol{x})_{h,w}$ is large, it indicates that the logits of the pixel across all inlier classes are small, \ie, the pixel $\boldsymbol{x}_{h,w}$ is dissimilar from all inlier classes. 

\subsection{RbA}
RbA uses Mask2Former (see Fig~\ref{fig:architecture}(b)), a mask-based network, as its backbone. 
Let's define $\boldsymbol{l}_{h,w}\left(k\right)$ is the class logit of a pixel $\boldsymbol{x}_{h,w}$ for class $k$. 
In Mask2Former model, the $\boldsymbol{l}_{h,w}\left(k\right)$ is described as: 

\begin{equation}
    \boldsymbol{l}_{h,w}\left(k\right) = \sum_{i=1}^N P_n(\boldsymbol{x_{h,w}};k) M_n(\boldsymbol{x_{h,w}})
    ,
  \label{eq:mask_logit}
\end{equation}
where $N$ is the total number of refined query in the network, $M_n(\boldsymbol{x_{h,w}})$ denotes the mask prediction score, generated by the $n$-th query, on the pixel $\boldsymbol{x_{h,w}}$, and $P_n(\boldsymbol{x_{h,w}};k)$ indicates the $k$-th class predicted probability of the $n$-th query. 
In other words, the class logit $\boldsymbol{l}_{h,w}\left(k\right)$ is the weighted sum of the classification predictions of each query, and the weights are the mask prediction scores generated by each query for the pixel $\boldsymbol{x_{h,w}}$. 
Based on the Eq~\ref{eq:mask_logit}, the anomaly scoring function $\textup{RbA}$ is defined as: 

\begin{equation}
    \textup{RbA}(\boldsymbol{x_{h,w}}) = -\sum_{k=1}^K \sigma (\boldsymbol{l}_{h,w}\left(k\right))
    ,
  \label{eq:rba1}
\end{equation}
where $K$ is the total number of inlier class. When the anomaly score has a large value, it indicates that all class logits, $\boldsymbol{l}_{h,w}\left(k\right)$ where $k\in\{1,\cdots,K\}$,  are small. This means that the mask scores generated by each refined query for this pixel are all small, indicating that the pixel is not similar to any inlier class. 

\subsection{Mask2Anomaly}
Similar to RbA, Mask2Anomaly also uses Mask2Former as its backbone. Therefore, its definition of the class logit $\boldsymbol{l}_{h,w}\left(k\right)$ is the same as Eq~\ref{eq:mask_logit}. 
Mask2Anomaly compute the anomaly score $f(\boldsymbol{x_{h,w}})$ as: 

\begin{equation}
    f(\boldsymbol{x_{h,w}}) = 1-\max_{k=1}^K \sigma (\boldsymbol{l}_{h,w}\left(k\right))
    ,
  \label{eq:rba2}
\end{equation}
When the value of $f(\boldsymbol{x_{h,w}})$ is large, it implies that the value of $\max_{k=1}^K \sigma (\boldsymbol{l}_{h,w}\left(k\right))$ is small, meaning that all class logits, $\boldsymbol{l}_{h,w}\left(k\right)$ where $k\in\{1,\cdots,K\}$, are small. 
This indicates that the mask scores generated by each refined query for this pixel are small, implying that the pixel is not similar to any inlier class. 

\section{Additional Results and Implementation Details}

In our main paper, we validate that the proposed RWPM can improve the performance of existing anomaly segmentation methods by integrating our approach with four recent representative methods (see Section 4.1 of the main paper). However, due to page limitations, we only present results on the Fishyscapes Lost \& Found dataset and the Road Anomaly dataset. Therefore, in Table~\ref{tab:s_effect}, we provide additional results, including on the Anomaly track and Obstacle track of the SMYIC dataset. From Table~\ref{tab:s_effect}, it demonstrates that our proposed RWPM consistently and significantly enhances the performance of existing anomaly segmentation methods across different datasets. 

Furthermore, in Table~\ref{tab:s_para}, we present the implementation details of the RWPM parameters $\alpha$ (the transition probability) and $T$ (the number of iteration) in each experiments in Table~\ref{tab:s_effect}. 

\begin{table}[htbp]
\tabcolsep=5pt
\caption{Comparison with strong and representative anomaly segmentation baselines across different backbone Architectures (Arch). \textbf{Bold} denotes the better results between using RWPM and not using RWPM with the same anomaly segmentation baselines. \textcolor{red}{$^\dagger$ } indicates that use the calibration in sub-map concatenation as described in Section 3.4 of the main paper.}
\centering
\resizebox{1.0\linewidth}{!}{
    \begin{tabular}{l|l|ccc|ccc|cc|cc}
    \toprule
    \multicolumn{2}{l}{Benchmark$\rightarrow$} & \multicolumn{3}{c}{Fishyscapes Lost\&Found} & \multicolumn{3}{c}{Road Anomaly} & \multicolumn{2}{c}{Anomaly Track}& \multicolumn{2}{c}{Obstacle Track} \\
    \midrule
    Method $\downarrow$ &Arch    & AuROC$\uparrow$          & AP$\uparrow$         & FPR95$\downarrow$         & AuROC$\uparrow$      & AP$\uparrow$      & FPR95$\downarrow$   & AP$\uparrow$      & FPR95$\downarrow$ & AP$\uparrow$      & FPR95$\downarrow$  \\
    \midrule
    PEBAL~\cite{tian2022pixel}  &Pixel-based        &98.96           &58.81       &4.77           &87.63       &45.10    &44.58         & 49.13 & 40.87 & 4.96 & 12.71  \\
    PEBAL + RWPM$\textcolor{red}{^\dagger}$ (Ours) &Pixel-based&\textbf{99.20}  &\textbf{66.85} & \textbf{3.68} &\textbf{89.48} &\textbf{50.29} & \textbf{36.81}   & \textbf{51.49} & \textbf{34.04} & \textbf{14.47} & \textbf{11.23}       \\
    \midrule
    Balanced Energy~\cite{choi2023balanced}  &Pixel-based   &99.03             & 67.07      & 2.93          & 88.31      &41.48    & 41.46   & 47.70 & 52.00 & 9.07 & 18.78  \\
Balanced Energy + RWPM$\textcolor{red}{^\dagger}$ (Ours) &Pixel-based & \textbf{99.29} & \textbf{72.95} & \textbf{2.38}  & \textbf{90.51} & \textbf{48.26}& \textbf{32.10}  & \textbf{50.44} & \textbf{51.13} & \textbf{20.00} & \textbf{16.65}     \\
    \midrule
    Mask2Anomaly\tablefootnote[1]{We carefully reproduce the experiments by using the official code.}~\cite{rai2023unmasking}    &Mask-based        & 95.47            & 65.27       &  7.79        & 96.54       & 80.04  &  13.95 & 88.62 & \textbf{14.57} & 93.10 & 0.20        \\
    Mask2Anomaly + RWPM (Ours) &Mask-based  &  \textbf{95.48}        &  \textbf{65.38}      & \textbf{7.33}  & \textbf{97.44} & \textbf{80.09} & \textbf{7.45}  & \textbf{88.98} & 15.88 & \textbf{93.82} & \textbf{0.18} \\
    \midrule
    RbA~\cite{nayal2023rba}    &Mask-based                  &  98.62         &   70.81    &   6.30        &  97.99     &  85.42  & 6.92    & 90.86 & 11.59 & 91.85 & 0.46   \\
    RbA + RWPM (Ours) &Mask-based      &  \textbf{98.82}      &   \textbf{71.16}      &  \textbf{6.12}           & \textbf{98.04} &  \textbf{87.34}  & \textbf{5.27}  & \textbf{92.00} & \textbf{10.15} & \textbf{93.30} & \textbf{0.28} \\
    \bottomrule
    \end{tabular}}
\label{tab:s_effect}
\end{table}

\begin{table}[htbp]
\tabcolsep=12pt
\caption{The implementation detail of RWMP parameters $\alpha$ (the transition probability) and $T$ (the number of iteration) for each experiment in in Table~\ref{tab:s_effect}. \textcolor{red}{$^\dagger$ } indicates that use the calibration in sub-map concatenation as described in Section 3.4 of the main paper.}
\centering
\resizebox{1.0\linewidth}{!}{
    \begin{tabular}{l|cc|cc|cc|cc}
    \toprule
    \multicolumn{1}{l}{Benchmark$\rightarrow$} & \multicolumn{2}{c}{Fishyscapes L\&F} & \multicolumn{2}{c}{Road Anomaly} & \multicolumn{2}{c}{Anomaly Track}& \multicolumn{2}{c}{Obstacle Track} \\
    \midrule
    Method $\downarrow$    & $\alpha$  & $T$   & $\alpha$  & $T$  & $\alpha$  & $T$  & $\alpha$  & $T$    \\
    \midrule
    PEBAL + RWPM$\textcolor{red}{^\dagger}$ (Ours) & 0.99 & 5 & 0.99 & 20 & 0.99 & 5 & 0.99 & 5\\
    Balanced Energy + RWPM$\textcolor{red}{^\dagger}$ (Ours) & 0.90 & 5 & 0.99 & 20 & 0.99 & 5 & 0.99 & 5\\
    Mask2Anomaly + RWPM (Ours) & 0.10 & 5 & 0.99 & 20 & 0.90 & 5 & 0.70 & 5  \\
    RbA + RWPM (Ours) & 0.50 & 5 & 0.99 & 20 & 0.99 & 5 & 0.90 & 5\\
    \bottomrule
    \end{tabular}}
\label{tab:s_para}
\end{table}

\section{Different Methods for Constructing Locally Constrained Graph}
In our main paper (see Section 3.2), we utilize a Softmax function to obtain locally constrained graph. 
In this section, we first present a comparison between using the Softmax function and employing $k$-NN search algorithm in our RWPM. 
When using the $k$-NN algorithm, for each pixel, we only retain its connections to the $k$ most similar pixels (neighborhoods) in the graph $\mathbf{S}$. 
Table~\ref{tab:softmax} shows the results. 
We observe that when using the $k$-NN algorithm, as the value of $k$ decreases (from $k=500$ to $k=50$), the performance improves due to the reduction of noise effect. 
This demonstrates the effectiveness of the locally constrained graph. 
However, when the value of $k$ becomes too small (from $k=50$ to $k=10$), it leads to the loss of relationship information between pixels, resulting in a decrease in performance. 
Nevertheless, although the $k$-NN algorithm can achieve good performance, it significantly reduces operational efficiency (with only $0.93$ FPS). In contrast, our proposed Softmax method can achieve better performance and higher operational efficiency (with $2.04$ FPS).
This demonstrates the superiority of our Softmax method. 

\begin{table}[htbp]
\tabcolsep=10pt
\caption{The comparison between the performance of Softmax functions and $k$-NN search algorithms using different $k$ values. 
We set $T = 20$ and $n = 2$ for all experiments. 
\textbf{Bold} and \underline{underline} denote the best and the second best results, respectively.}
\centering
\resizebox{0.6\linewidth}{!}{
    \begin{tabular}{lccc}
    \toprule
    \multicolumn{1}{l}{Benchmark$\rightarrow$} & \multicolumn{3}{c}{Road Anomaly}  \\
    \midrule
     & \textbf{AP$\uparrow$}    & \textbf{FPR95$\downarrow$}  & \textbf{FPS$\uparrow$}  \\
    \midrule
    $k$-NN ($k=10$)  &86.38 &6.57 &0.93    \\
    $k$-NN ($k=20$)  &86.65 &6.47 &0.93   \\
    $k$-NN ($k=50$)  &\underline{86.71} &\underline{5.96} &0.93   \\
    $k$-NN ($k=100$)  &86.24 &6.00 &0.93    \\
    $k$-NN ($k=200$)  &84.61 &6.72 &0.93    \\
    $k$-NN ($k=500$)  &81.00 &19.08 &0.93    \\
    \midrule
    Softmax ($\tau = 0.01$)  &\textbf{87.34} &\textbf{5.27} &\textbf{2.04}  \\
    
    \bottomrule
    \end{tabular}}
\label{tab:softmax}
\end{table}

Additionally, we show the effect of the temperature parameter $\tau$ in the Softmax function. 
A smaller $\tau$ results in fewer neighborhoods being connected to each pixel in the graph $\mathbf{S}$, which resembles the reduction of the $k$ value in $k$-NN. 
Table~\ref{tab:tau} presents the results. 
We can observe a similar phenomenon to that in Table~\ref{tab:softmax}, where the locally constrained graph can improve the performance, but when too few pixels are connected to each pixel, it may actually decrease the performance. 
In our experiments, the best results were achieved when $\tau=0.01$.

\begin{table}[htbp]
\tabcolsep=15pt
\caption{The effect of temperature parameter $\tau$. 
We set $n=2$ and $T=20$ for all experiments. 
\textbf{Bold} denotes the best results.}
\centering
\resizebox{0.5\linewidth}{!}{
    \begin{tabular}{lcccc}
    \toprule
    \multicolumn{1}{l}{Benchmark$\rightarrow$} & \multicolumn{2}{c}{Road Anomaly}  \\
    \midrule
     & \textbf{AP$\uparrow$}    & \textbf{FPR95$\downarrow$}   \\
    \midrule
    RbA w/o RWPM  &85.42 &6.92   \\
    \midrule
    $\tau = 1.0$   &21.60 &79.25   \\
    $\tau = 0.1$   &31.45 &68.65   \\
    $\tau = 0.01$   &\textbf{87.34} &\textbf{5.27}   \\
    $\tau = 0.001$   &85.97 &6.70   \\
    $\tau = 0.0001$   &85.57 &6.93   \\
    \bottomrule
    \end{tabular}}
\label{tab:tau}
\end{table}

\section{Analysis for RWPM}

\begin{figure}[htbp]
    \centering
    \begin{tabular}{ccc}
        \includegraphics[width=0.28\linewidth]{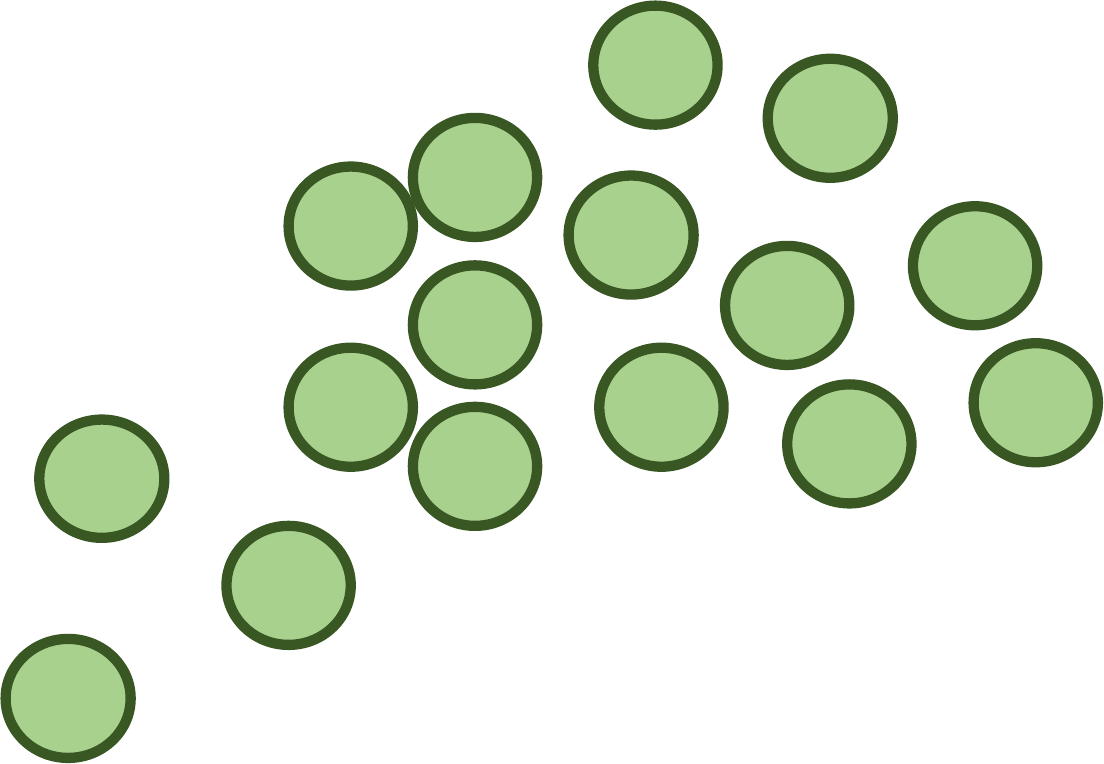} &
	\includegraphics[width=0.28\linewidth]{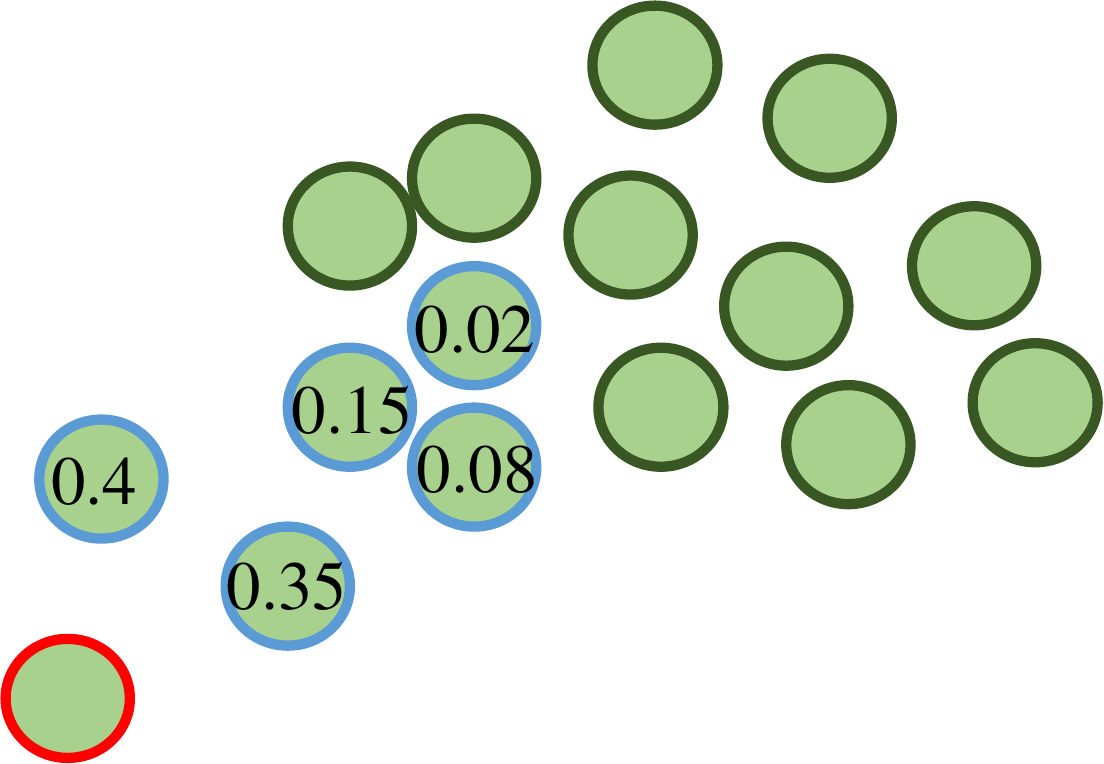} &
	\includegraphics[width=0.28\linewidth]{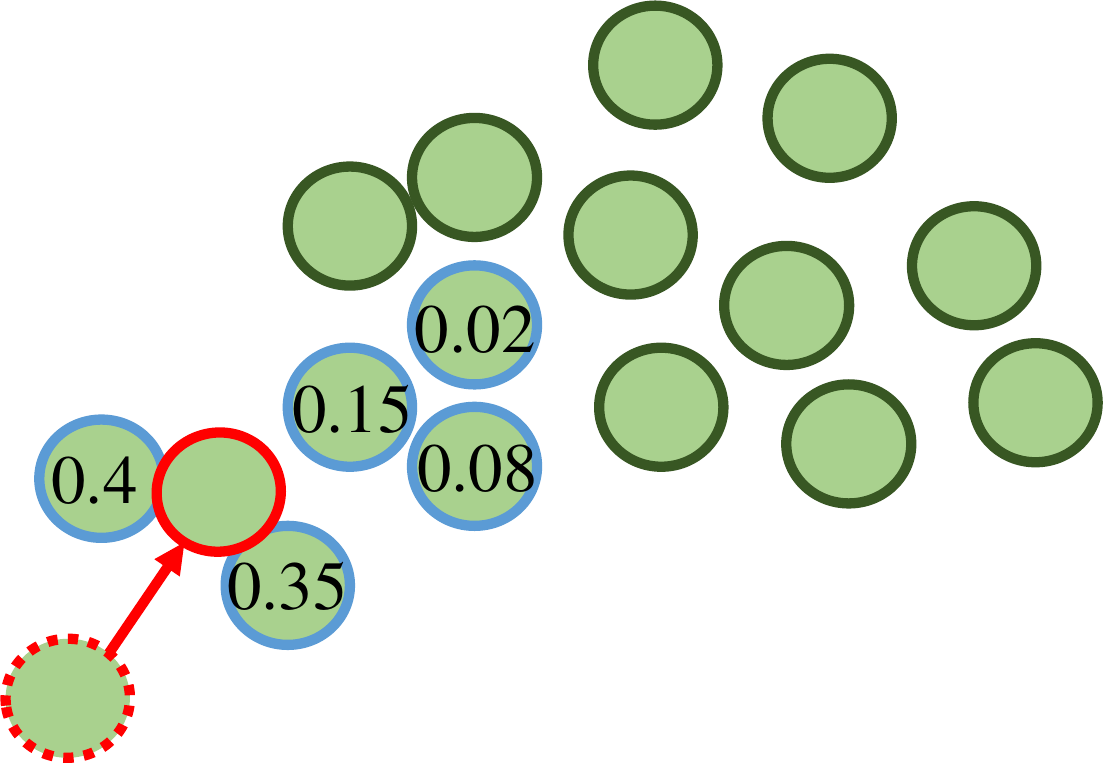} \\
    (a) Pixel Distribution & (b) Pixel and Its Neighborhoods & (c) Pixel Embedding Updating
    \end{tabular}
    \caption{\textbf{The process of updating pixel embeddings.} (a) shows the distribution of pixel embeddings, all data points belong to the same category. (b) shows a pixel embedding (red circle) that deviates from the clustering and its $5$ neighborhoods (blue circle). The number inside each neighborhood (blue circle) represents its similarity to the pixel embedding (red circle). (c) shows that, after the computation by the weighted ensemble, the updated pixel embedding moves closer to the clustering. }
    \label{fig:rw_process}
\end{figure}

In this section, we analyze why RWPM can optimize pixel embeddings.
In the main paper, the process of RWPM can be described as: 

\begin{equation}
    \begin{matrix}
    \mathbf{m}^{t+1} = \alpha \mathbf{S} \mathbf{m}^{t}+\left(1-\alpha\right)\mathbf{m}^{0}, & \alpha \in \left(0,1\right)
    \end{matrix}
    ,
    \label{eq:random_walk_supp}
\end{equation}
where $\mathbf{S}$ denotes the graph representing the manifolds of pixel embeddings, $\mathbf{m}^{0}$ denotes the original pixel embeddings, $\mathbf{m}^{t}$ denotes the updated pixel embeddings in the $t$-th iteration of the random walks and $\alpha$ denotes the continuing probability. 
Eq~\ref{eq:random_walk_supp} shows that each updated pixel embedding is obtained by taking a weighted ensemble of other pixel embeddings. The weights are determined by the similarities between the pixels depicted on graph $\mathbf{S}$. 
As shown in Fig~\ref{fig:rw_process}, after updating, each pixel will become more similar with its neighborhoods. In other words, all pixel embeddings in the same manifolds will become more similar and form a more compact cluster. 

Additionally, similar to the derivation in ~\cite{bai2017regularized}, the closed-form solution of Eq~\ref{eq:random_walk_supp} can be formulated as the following optimization problem: 

\begin{equation}
    \min_{\mathbf{m}^{\infty}} \frac{1}{2} \sum_{i,j=1}^{HW} \mathbf{S}_{ij}\left(\mathbf{m}^{\infty}_i - \mathbf{m}^{\infty}_j \right)+\frac{1-\alpha}{\alpha} \sum_{i=1}^{HW} \left(\mathbf{m}^{\infty}_i - \mathbf{m}^{o}_i \right)^2
    ,
    \label{eq:random_walk_optim}
\end{equation}
where $\mathbf{m}^{\infty}$ is the finial updated embeddings, $\mathbf{m}_{i}$ denotes the embedding of the $i$-th pixel and $\mathbf{S}_{ij}$ depicts the similarity between the $i$-th and $j$-th pixels on the manifolds. 
We can observe that Eq~\ref{eq:random_walk_optim} consists of two terms: 
The first term indicates that if two pixels are more similar on manifolds, their final embeddings will also be more similar. 
In other words, the first term considers higher-order information among pixels.
For example, if the $i$-th pixel is similar to the $j$-th pixel, and the $j$-th pixel is similar to the $k$-th pixel, then in the final state, the embedding of the $i$-th pixel will also be similar to that of the $k$-th pixel. 
These encourage pixels on the same manifold to form more compact clusters, consistent with the description and experimental results in the main paper.
The second term suggests that the initial embeddings should be retain to a certain degree. 
In our limited iteration strategy, $\mathbf{m}^{t}$ is regarded as an approximation of $\mathbf{m}^{\infty}$.

% \begin{equation}
%     \min_{\mathbf{m}^{\infty}} \frac{1}{2} \sum_{i,j=1}^{N} \mathbf{S}_{ij}\left(\mathbf{m}^{\infty}_i - \mathbf{m}^{\infty}_j \right)+\frac{1-\alpha}{\alpha} \sum_{i=1}^{N} \left(\mathbf{m}^{\infty}_i - \mathbf{m}^{o}_i \right)^2
%     ,
%     \label{eq:random_walk_optim}
% \end{equation}
% where $\mathbf{m}^{\infty}$ indicates the final refined pixel embeddings, $\mathbf{m}_i$ indicates the $i$-th pixel embedding and $\mathbf{S}_{ij}$ is the similarity between the $i$-th and $j$-th pixel described by graph $\mathbf{S}$. 
% Eq~\ref{eq:random_walk_optim} shows that 

\section{Some Reviewers' Comments and Our Explanations}

\noindent \textbf{Comment\#1 [Why are random walks particularly effective for anomaly segmentation?]:} 

\noindent Existing anomaly score functions rely on the logits output by segmentation models to infer anomaly scores. However, diverse real-world driving scenarios often distort manifolds of pixel features 
(but most pixels of same class still lie on the same manifold), 
leading to inaccurate anomaly score inference (see Fig. 1 of the main paper). Thus, we employ random walks on manifolds to propagate and update pixel features. Random walks can measure the similarity of pixels on manifolds (\ie, distance on manifolds rather than Euclidean distance), resulting in high similarity for pixels on the same manifold and low similarity for pixels on different manifolds. This forms more compact clusters on each manifold, mitigating manifold distortion. This is why random walks are effective for anomaly segmentation. 

\noindent \textbf{Comment\#2 [The proposed partitioning strategy sounds reasonable, but what effect does it have when anomaly parts are split among different sub-maps? Did the authors observe this as an issue?]:}  

\noindent From the analysis of numerous samples, we observe that when the number of sub-maps $n$ is small, the high similarity between sub-maps contents ensures that dividing the anomaly target into separate parts does not affect the results. However, when $n$ is larger, our proposed calibration method can alleviate the issue (see Section 3.4 of the main paper). Naturally, finding more elegant solutions to this problem will be part of our future work. 

\noindent \textbf{Comment\#3 [It would be interesting to know RWPM’s limitations and which solutions could be adopted to alleviate/solve such limitations.]:} 

\noindent One limitation of RWPM is running efficiency. 
Although we have proposed Partial Random Walk and verified its effectiveness in improving efficiency, there are still some methods that can further enhance operational efficiency in practical applications. For instance, in practical applications, we can use image segmentation results to remove background elements (\eg, sky) or other irrelevant regions. This significantly reduces the pixel count and improves computational efficiency, helping to meet real-time requirements. Besides, as we mentioned in the conclusions section, another limitation is that RWPM's effectiveness depends on the model's quality. Training models with RWPM to enhance generalization is a promising solution. 

% \bibliography{supp.bib}

\end{document}